\documentclass[twocolumn]{article}
\pdfoutput=1 
\usepackage[utf8]{inputenc}
\usepackage{subcaption}
\usepackage{graphicx}
\usepackage{threeparttable}
\usepackage{amsmath, bm}
\usepackage[dvipsnames]{xcolor}
\usepackage{hyperref}
\usepackage{cleveref}
\usepackage[noend]{algorithmic}
\usepackage{algorithm}

\newcommand\snr{SNR optimality}

\newcommand\nl{NL-SGD}
\newcommand\nlm{NL-Mom}
\newcommand\nln{NL-NAG}
\newcommand\sign{\text{sgn}}
\newcommand{\Loss}{\mathcal{L}}
\newcommand{\demph}{\textcolor{gray}}
\newcommand\myshade{90}
\colorlet{mylinkcolor}{NavyBlue}
\colorlet{mycitecolor}{Aquamarine}
\colorlet{myurlcolor}{Aquamarine}

\hypersetup{
  linkcolor  = mylinkcolor!\myshade!black,
  citecolor  = mycitecolor!\myshade!black,
  urlcolor   = myurlcolor!\myshade!black,
  colorlinks = true,
}

\usepackage[margin=2.6cm]{geometry}
\usepackage{libertine}
\usepackage{libertinust1math}
\usepackage{inconsolata}
\usepackage[T1]{fontenc}    
\usepackage[sf,bf]{titlesec}

\usepackage{parskip}

\usepackage[backend=biber,
style=numeric,
sorting=none,
isbn=false,
doi=false,
url=false,
]{biblatex}

\title{\textsf{\textbf{On the benefits of non-linear weight updates}}}
\author{
  Paul Norridge\footnote{
  \texttt{\small paul.norridge@gmail.com}}
  }

\begin{document}

\maketitle
\begin{abstract}
    Recent work has suggested that the generalisation performance of a DNN is related to the extent to which the Signal-to-Noise ratio is optimised at each of the nodes. In contrast, Gradient Descent methods do not always lead to SNR-optimal weight configurations. One way to improve SNR performance is to suppress large weight updates and amplify small weight updates. Such balancing is already implicit in some common optimizers, but we propose an approach that makes this explicit. The method applies a non-linear function to gradients prior to making DNN parameter updates. We investigate the performance with such non-linear approaches. The result is an adaptation to existing optimizers that improves performance for many problem types.
\end{abstract}

\section{Introduction}
Deep neural networks are being used ever more widely for machine learning applications and continue to demonstrate impressive results. Critical to this success has been the widespread use of back-propagation \cite{backprop} via Gradient Descent (and, more usually, Stochastic Gradient Descent, SGD). Despite the great success of deep learning, the precise reasons why some networks generalise better than others is not always clear \cite{Chiyuan_Zhang_et_al_2016, chiyuan_still}.\par 

Recently it has been proposed that a contributor to the generalisation performance of DNNs is the `\snr{}' of the weight set employed, see \cite{snr}. That is, for a given data set, do the weights act to optimise the Signal-to-Noise ratio (SNR) at the output of the nodes in the network? In particular, when a node has correlated inputs, do the weights maximise the use of all these inputs to combat noise? Empirical evidence suggests that this is indeed a contributor to performance. \par

This result leads to the question of application and whether it is possible to find training methods that enhance the \snr{} of a network. In this paper, we investigate one such SNR-inspired enhancement to training. The purpose of this is twofold: First, to extend the study in \cite{snr}, by developing an optimization approach inspired by the \snr{} perspective and investigating its effectiveness; Second, to propose  adaptations to standard optimizers that can be used to improve generalisation.\par

We emphasise that we do not intend to formally prove the validity of the optimizer adaptations proposed. The approach is to consider the weaknesses of optimisers such as SGD and Momentum from an \snr{} point-of-view and identify ways that these might be mitigated. The benefit of the approach is shown by applying the adapted optimizers to a set of test benches.\par

The paper is structured as follows: We begin by giving some details on the SNR motivation for the approach and outlining the geometric picture that this viewpoint suggests. We then describe the key features of the proposed non-linear adaptation and the specific implementations considered here. These implementations are then tested against a set of benchmarks. The appendices provide some additional `toy model' motivation and plots of the training evolution.\par

\section{Related Work}

The work presented here overlaps with three areas of Machine Learning study.\par

The main focus is the development and adaptation of optimizers for Gradient Descent training of DNNs.  A significant number of optimizers have been proposed in an effort to ensure that DNN training leads to networks that perform well, while minimising training time. It is not possible to list all of the options here, but we use a number as reference points; basic Stochastic Gradient Descent \cite{Robbins1951}, Momentum \cite{Polyak1964}, Nesterov Accelerated Gradient (NAG)  \cite{Nesterov1983}, Adam  \cite{Kingma2015}, AMSGrad  \cite{Reddi2018}, NADAM \cite{Dozat2016IncorporatingNM}, Lookahead Momentum \cite{Zhang2019a}, AMS Bound \cite{Luo2019} and RMSProp \cite{Tieleman2012}. In addition, \cite{powersign} is an interest comparison, since it uses non-linear functions of the gradient, but only to alter the effective learning rate. Our benchmark is Deepobs \cite{deepobs} and in particular the survey provided in \cite{crowdedvalley}. We also note the useful overviews provided in \cite{Ruder16} and \cite{DLrefined}.\par

Close in spirit to the work here is gradient clipping \cite{difficultytraininglanguage, Mikolov}. Although this is commonly implemented as a distinct step in training calculations, it can be thought of as an adaptation applied to optimizers. (In fact, this is how it is treated in Keras \cite{chollet2015keras}.)\par

To a lesser extent, we will also touch on the question of the relationship between flat minima and generalisation performance. This relationship has been an area of investigation since the original observations in  \cite{Sepp_Hochreiter_Jrgen_Schmidhuber1997}. Again, it is not possible to list the full list of papers discussing this area, but we note the recent discussions in \cite{Nitish_Shirish_Keskar_et_al_2016},  \cite{Laurent_Dinh_et_al_2017}, \cite{ExploringGeneralization}, \cite{2017arXiv171005468K} and \cite{flatminima_largemargins}. Of particular interest here is \cite{asymmetric_flatminima}, which shows that minima may only have a subset of directions that are flat. This is consistent with the expectations motivating this study.  Recently, there has been interest in optimization approaches that favour flat minima, see, for example, \cite{flatminima_comparison}, \cite{EntropySGD} and \cite{sharpness_aware}. The investigation here is primarily concerned with \emph{where} in a minima the model is left after optimization rather than explicitly looking for such regions. \par

In addition, the work here is motivated by noise robustness. Although we build primarily on the investigation in \cite{snr}, we note that noise robustness has also been used in generalisation predictions, as described in \cite{fantastic_measures}, \cite{predicting_generalisation_via_noise}and \cite{natekar2020representation}.\par

Tying the last two points together, in some sense, is ref \cite{uniqueproperties_flatminima}, which demonstrates that flat minima have paths through the network that are dedicated to signals with the highest gain. \par

\section{Motivation}\label{sec.motivation}

\subsection{\snr{}}

The motivation for the approach we will propose is to select weight sets that minimise the impact of noise on the network performance, or conversely, optimise the preservation of useful information. We can summarise the reasoning as follows:
\begin{enumerate}
\item{A DNN will perform best when the available (useful) information is exploited to maximum extent.}
\item{Optimal use of information by the entire network depends on maximising the information preservation of individual nodes.}
\item{The information preservation from input to output of a node can be characterised via the Signal-to-Noise Ratio (SNR)}
\item{It is possible to identify an optimal set of weights to maximise SNR, which motivates the way we train the network.}
\end{enumerate}

Here we briefly recap the results from \cite{snr}. \par

Our building block is the usual node definition 
\begin{equation} 
y = \sum_jw_{ij}x_{j} + b_i
\end{equation}
with inputs $x_j$, weights $w_{ij}$, bias $b_i$ and activation function $g()$.\par

We consider that every input has a contribution that is useful information -- signal -- together with noise. Partitioning the inputs into signal and noise components, the weighted sum can be expressed as:
\begin{equation} 
\sum_jw_{ij}x_{j}  = \sum_jw_{ij}\left(s_j + n_j\right) 
\end{equation}

Initially, we leave open the question of how to distinguish `signal' and `noise'. But, critically, we assume that the signal is a component that is shared by some subset of the inputs; that is, inputs have correlated components.\par

The SNR for the node can be expressed as \cite{R._Linsker1988}
\begin{equation}\label{eq:snr}
SNR_i^{(m)} = \frac{\text{var} \left(\sum_{j}w_{ij}s_{j} \right)}{ \sum_{j}w_{ij}^2\text{var} \left( n_{j} \right) } 
\end{equation}
where we have assumed that the noise components from different inputs are statistically independent. \par

Maximising this expression, we obtain a term for the optimal weight set.

\begin{equation}
w_{ij}= k_i .  \frac{\text{cov} \left( s_j, \sum w_{ik} s_k \right) }{\text{var} \left( n_j \right)} 
\label{eq:optimal_wts}
\end{equation}
where $k_i$ are contributions common to all inputs to the node.\par

We note that we are essentially requiring that the weights form a matched filter (see also \cite{matchedfilter_perspective} for a matched filter approach to DNNs\footnote{Not discussed in \cite{snr}  or \cite{matchedfilter_perspective} is that the activation function can be characterised as a matched filter detection, with the node bias being the detection threshold.}). \par

In \cite{snr}, this was taken further by assuming that the signal can be defined by the output of the node. That is, that the training process causes the output to be dominated by useful information. This led to an expression for the optimal weight set that was shown to be correlated to the generalisation performance of a number of examples. Based on the results presented, there is evidence that a DNN will generalise well when it has good performance on the training dataset \emph{and} the weights are close to this optimum. \par

With this in mind, a more intuitive way to interpret equation \ref{eq:optimal_wts} is that, after convergence, for each node, we want any inputs that have good correlation with the output to have corresponding high weighting.\par


However, we do not need to make assumptions about the content of the signal for the current discussion; if we accept that SNR maximisation is beneficial, we can use this matched filtering expression (equation  \ref{eq:optimal_wts}),  to identify possible weaknesses of SGD and other optimizers.\par

First, observe that even if two inputs are correlated, they will typically be treated differently to each other by the training process. There is an aspect of success-breeds-success in the SGD process. That is, if there are two inputs to a node with the same SNR but different signal levels, the higher magnitude input may be enhanced the most, even though noise sensitivity would be improved if the two had equal emphasis. In a DNN, this can lead to feedback where the higher magnitude inputs get increasingly amplified throughout the network at the cost of performance; a `toy model' demonstrating this is given in appendix \ref{sec.toy}.\par

Second, if there are different activation rates on inputs to a node, this can cause the weighting to be applied sub-optimally. Consider the following:
If a ReLU activation has been applied to the inputs, $x_i \leftarrow \text{max}(0,x_i)$, noise will only contribute when the $x_i>0$. That is, the noise will have zero variance whenever $x_i<0$. To make this explicit, we re-express the noise variance as
\begin{equation}
 \text{var} \left( n_{j} \right) \approx a_j ~ \text{var}(n^r_j)
\end{equation}

where  $a_j$ is the rate of activations of node $j$ of preceding layer, calculated as  $a_i = p(x_i > 0)$, and $\text{var}(n^r_j)$ is the `raw' noise variance of samples prior to application of the activation function.\par

\begin{equation}
w_{ij}= k_i .  \frac{\text{cov} \left( s_j, \sum w_{ik} s_k \right) }{ a_j ~ \text{var}(n^r_j)}  
\label{eq:optimalwts_withactivation}
\end{equation}

So, the optimal SNR configuration depends on the activations of the connected input nodes.\par

Hence, \snr{} requires that we compensate for the activation rate of an input when selecting optimal weights. We should select weights based on the contribution of the inputs when non-zero rather than their contribution in all cases. Gradient Descent typically does not take this into account -- a node with higher activation rate will normally have larger gradients and be emphasised more. As a consequence, inputs with relatively low activation rates will typically have lower weighting than would be optimal from an \snr{} point-of-view.\par

Of course, the reduction of the difference between large and small weight updates is an implicit feature of many optimizers, but typically this is a by-product of a design which is aimed at other considerations. For example, a valuable characteristic attributed to Momentum is that it allows better balance between directions with large and small gradients (see, for example, \cite{goh2017why, JMLR:v17:15-084}). Optimizers with normalisation, such as Adam, have similar benefits. Ref. \cite{adam} notes that the Adam weight updates are scale invariant (so, will not depend on the weighting of the input or the activation rate) and the magnitude of the updates are related to some extend to the noise on the inputs. \par

Our interest here is whether we can find an effective optimizer that takes inspiration directly from equation (\ref{eq:optimal_wts}) rather than improving \snr{} as a by-product.\par

\subsection{Geometric intuition}

It is beneficial to have a picture of the error surface that the \snr{} perspective implies. Intuitively, if a node has correlated inputs then we expect to find it associated with a minimum that is almost flat, in the sense that some directions will have small gradients. In the ideal case, where a subset of inputs are perfectly correlated and the same magnitude, these inputs will be interchangeable. Consequently, changing the weighting over the subset will have a no effect so long as the sum of the weights remains the same. \par
That is, if $C$ is the set such that if $s_j = s ~ \forall j \in C$
\begin{align*}
    \sum_jw_{ij}x_{j}  &= \sum_{j \in C}w_{ij}s_j + \sum_{j \notin C}w_{ij}x_{j} &\\
    & = s\sum_{j \in C}w_{ij} + \sum_{j \notin C}w_{ij}x_{j}& 
\end{align*} 

In the case where the inputs also have noise components, the situation is similar, but some points on flat region will have more noise resilience than others -- to make it explicit for this example, if all inputs have IID Gaussian noise, the total noise goes like $\sqrt(\sum_{j \in C}w^2_{ij})$ whereas signal depends on $\sum_{j \in C}w_{ij}$. Consequently, equal weights would be more noise resilient than other combinations with the same signal output.  If the correlation is only approximate the situation is less straightforward but the principle holds.\par

On the other hand, we note that if correlated inputs are distributed across multiple nodes, then we may not find such a simple noise-resilient structure, since the signals can become non-trivially linked to other unrelated features. The linking between unrelated features is also likely to cause sub-optimal SNR performance. This suggests a reason why flat minima are frequently related to good generalisation performance.\par

Of course, we find here a connection with the extensive studies on the relationship between flat minima and good generalisation. In particular, we note ref \cite{asymmetric_flatminima}, which shows that flat minima with good generalisation  may only be flat only in a subset of directions. In the perspective considered here,  not all inputs will be correlated, in general.

If this picture is correct, then we can identify a critical problem: there is a specific region on the flat minimum that is optimal for \snr{}; however, once we reach the minimum, it becomes harder to move towards that optimal point. Indeed, optimizers such as Adam will reduce the weight updates significantly when the flat region is reached since the second moments of the updates will be significantly larger than the corresponding first moments. \par

If our goal is to get close to an SNR-optimal location, then we would like the training process to be close at the point where it first reaches the flat region; this will minimise the need for subsequent movement in flat directions. Since, we know the criteria for \snr{} weights in equation \ref{eq:optimalwts_withactivation}, we can identify potential strategies to meet this requirement. Whereas the motivation for other optimizers has frequently been concern over how training behaves on reaching minima or saddle points, we suggest that what happens on the way to a zero-curvature region is also significant. \par


\section{Non-linear weight updates}
Above we identified two shortcomings of SGD from the perspective of \snr{}. During the training process, both of these shortcomings are realised by some signal-carrying inputs being under-emphasised with relative small weight updates. This suggests that the training process would be improved if we can adjust the weight updates on the `useful' inputs so that smaller updates are scaled-up relative to larger ones. Consequently, we propose an adaptation of SGD that applies weight updates of the form
\begin{equation}
    \Delta w \sim h\left(\frac{\partial f}{\partial w}\right)
\end{equation}
where $h()$ is some non-linear function with 
\begin{enumerate}
    \item{$0\leq\frac{\partial h\left(x + \delta\right)}{\partial w}\leq\frac{\partial h\left(x\right)}{\partial w}\ \ \forall x, \delta > 0$}\\
    with at least one pair $x_1, \delta_1$  s.t. $ \frac{\partial h\left(x_1 + \delta_1\right)}{\partial w}<\frac{\partial h\left(x_1\right)}{\partial w}$
    \item{$\sign{\left(h(x)\right)} = \sign{\left(x\right)}$} 
    \item{h(x) = -h(-x)}
\end{enumerate} \par

Property 1 ensures that the function explicitly adjusts the SGD weight updates with an attenuation that is increased as the update magnitude increases. Hence, it effectively reduces the gap between large and small updates. \par

Stochastic Gradient Descent (as well as other optimizer formulations) has demonstrated that we don't need to consistently take the \emph{steepest} descent step at every training step; what is important is to continually head downhill. Property 2 ensures that this is the case.\par

Finally, property 3 ensures that there is no bias introduced. For example, if this was not respected, then uncorrelated inputs may grow purely from the stochastic gradients coming from mini-batches.\par 

We note that this approach is already implemented implicitly in some of the common weight update algorithms. For example, gradient clipping \cite{difficultytraininglanguage, Mikolov} fits this definition by applying 
\begin{equation}
    h\left(x\right) = \begin{cases}
x \text{\ for\ } x < t,\\
t \text{\ for\ } x > t\\
\end{cases}
\end{equation}

Similarly, component-wise normalisation (see, for example, \cite{mlrefined}) corresponds to an application of the function $h(x) = sgn(x)$. \par

Both of these options have weaknesses. Gradient clipping only restricts updates above a certain scale, which must be selected explicitly, prior to training; this limits the ability to balance updates. On the other hand, from an \snr{} point-of-view, $h(x) = sgn(x)$ risks doing too much; it treats all inputs equally, no matter how small the updates; this raises the possibility of introducing new contributions to a node `signal' and coupling unrelated features. We would like some middle-ground between the extremes of pure SGD and component-wise normalisation.  \par

In this paper, we propose an alternative function choice, 
\begin{equation}
    h_{\nu}\left(x\right) =\sign\left(x\right)\left|x\right|^\nu
\end{equation}
That is, with weight updates of the form
\begin{align*}
        \Delta w_{ij} &= \alpha\ h_{\nu}\left(\frac{\partial f}{\partial w_{ij}}\right)\\ 
        &=\alpha \ \sign{\left(\frac{\partial f}{\partial w_{ij}}\right)\left|\frac{\partial f}{\partial w_{ij}}\right|^\nu}
\end{align*}
where the exponent $\nu \in \left[0, 1\right)$ and $\alpha\in R^{>0}$ are both hyperparameters\footnote{We note that a variant on the $\nu = 1/2$ case was used in \cite{pohlen2018observe} and \cite{muzero} for the same kind of scaling envisaged here, albeit for scaling reinforcement learning value functions rather than gradients.\par }. In contrast to the simpler non-linear functions considered above, this has a scale-invarient effect, but does not give too much weight to very low gradients.\par

Clearly, at the endpoints of the interval for $\nu$, the weight update reduces to simple Gradient Descent and component-wise normalisation of the gradient; so, in some sense, this provides a continuous set of functions between these two extremes. \par

\begin{figure}[!htbp]
     \centering

       \includegraphics[width=.8\linewidth]{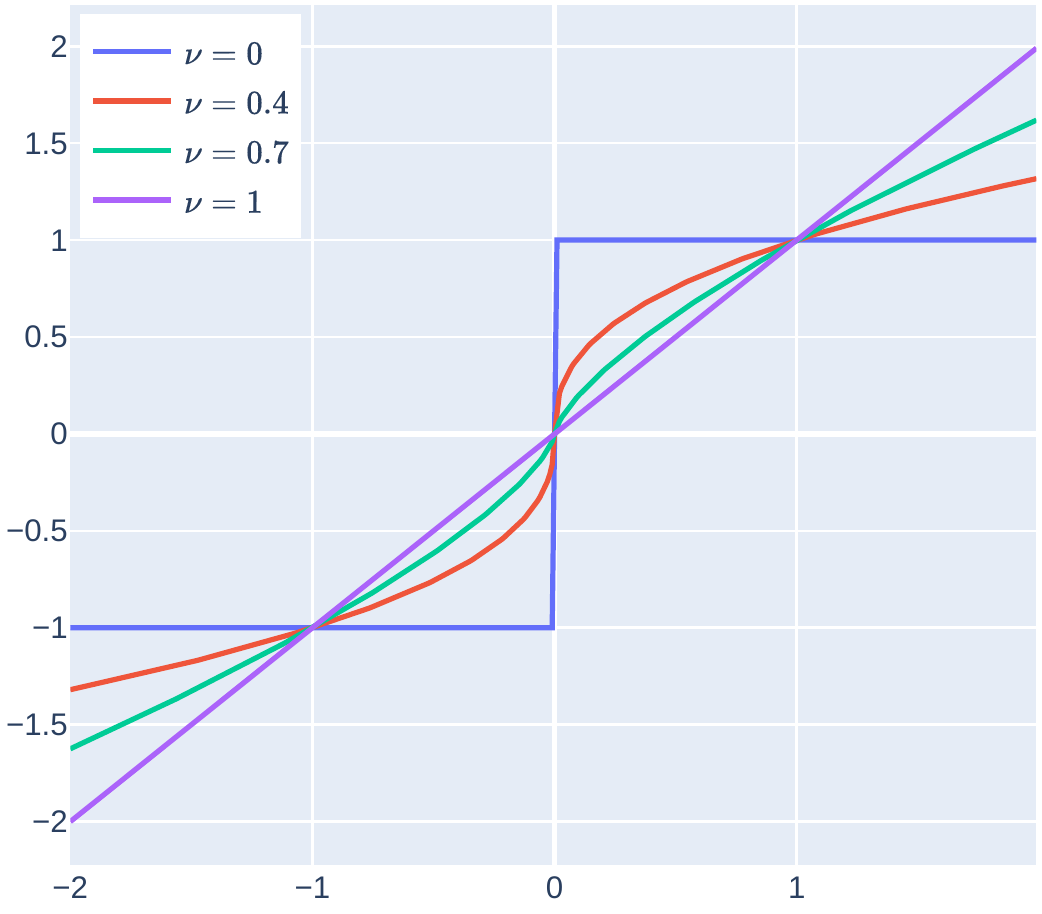}
       \caption{Example curves for $h_\nu$ functions}

     \label{fig:curves}
   \end{figure}

For reference, examples of the functions are given in figure \ref{fig:curves}.

Henceforth in this paper, we will label optimization algorithms using one of the $h_{\nu}$ functions with `NL-'. We detail some algorithm options below.\par

\begin{algorithm}[tb!]
\caption{NL-SGD with L$_2$ regularization} 
\footnotesize
\label{alg:nl}
\begin{algorithmic}[]
\STATE{\textbf{select} $\alpha \in \mathbb{R}^{+}, \nu \in \left(0,1\right], \lambda\in \mathbb{R}^+_0$}

\REPEAT
	\STATE{$t \leftarrow t + 1$}
	\smallskip
	\STATE{$\bm{g}^{\bm{\theta}}_t \leftarrow \frac{1}{m}\nabla_{\bm{\theta}} \sum_i L\left(f(x^{(i)};\bm{\theta}_{t-1}),y^{(i)}\right)$  \hfill  \text{ kernel gradients}  }
	
	\STATE{$\bm{g}^b_t \leftarrow \frac{1}{m}\nabla_{\bm{b}} \sum_i L\left(f(x^{(i)};\bm{b}_{t-1}),y^{(i)}\right)$  \hfill  \text{ bias gradients} }
	
	\medskip
	
	\STATE{$\bm{\theta} \leftarrow \bm{\theta}_{t-1} - \alpha h\left(\bm{g}^{\bm{\theta}}_t\right)  + \alpha\lambda\bm{\theta}_{t-1} $} 
	
	{$\bm{\theta} \leftarrow \bm{b}_{t-1} - \alpha \bm{g}^b_t  + \alpha\lambda\bm{b}_{t-1} $} 
	\medskip
\UNTIL{ \textit{stopping criterion is met}}

\end{algorithmic}
\end{algorithm}

\subsection{Regularisation}

When applying a NL function to the weight updates, we must consider how to introduce $L_2$ regularisation. It is clear that including an $L_2$ regularisation term directly in the loss function with our proposed update function will lead to changing regularisation behaviour depending on the magnitude of the primary loss term. \par

To address this issue, we follow the approach given in \cite{2017arXiv171105101L}. That work focussed on the appropriate way to apply weight decay in the context of the Adam optimizer, recommending that it is applied as a separate term in the weight update -- i.e. as explicit weight decay -- rather than including $L_2 = \lambda\|w\|^2$ in the loss function. \par

With this in mind, we propose the basic NL approach in algorithm  \ref{alg:nl}.

\subsection{Momentum}

As well as simple \nl{}, we will also consider Momentum and Nesterov Accelerated Gradient versions. These are applied simply by feeding our new weight update into the normal algorithm definitions. Although the adaptation is straightforward, for clarity we provide the descriptions as algorithms \ref{alg:nlmom} and \ref{alg:nlnag}. \par

\begin{algorithm}[tb!]
\caption{NL-Mom with weight decay} 
\footnotesize
\label{alg:nlmom}
\begin{algorithmic}[]
\STATE{\textbf{select} $\alpha, \rho \in \mathbb{R}^{+}, \nu \in \left(0,1\right], \lambda\in \mathbb{R}^+_0$} 

\REPEAT
	\STATE{$t \leftarrow t + 1$}
	\smallskip
	\STATE{$\bm{g}^{\bm{\theta}}_t \leftarrow \frac{1}{m} \nabla_{\bm{\theta}} \sum_i L\left(f(x^{(i)};\bm{\theta}_{t-1}),y^{(i)}\right)$  \hfill  \text{ kernel gradients}  }
	\STATE{$\bm{g}^b_t \leftarrow \frac{1}{m}\nabla_{\bm{b}} \sum_i L\left(f(x^{(i)};\bm{b}_{t-1}),y^{(i)}\right)$  \hfill  \text{ bias gradients} }
	\medskip
	
	\STATE{$\bm{v^{\bm{\theta}}_t} \leftarrow \rho\bm{v^{\bm{\theta}}_t} +  h\left(\bm{g}^{\bm{\theta}}_t\right)$}
	
	\STATE{$\bm{v^{b}_t} \leftarrow \rho\bm{v^{b}_t} + \bm{g}^{\bm{\theta}}_t$}
	\medskip
	
	\STATE{$\bm{\theta}_t \leftarrow \bm{\theta}_{t-1} - \alpha \bm{v^{\bm{\theta}}_t}   - \lambda\bm{\theta}_{t-1} $} 
	
	\STATE{$\bm{\theta}_t \leftarrow \bm{b}_{t-1} - \alpha\bm{v^{b}_t}   - \lambda\bm{b}_{t-1} $} 
	
	\medskip
\UNTIL{ \textit{stopping criterion is met}}

\end{algorithmic}
\end{algorithm}

\begin{algorithm}[tb!]
\caption{NL-NAG with weight decay} 
\footnotesize
\label{alg:nlnag}
\begin{algorithmic}[]
\STATE{\textbf{select} $\alpha, \rho \in \mathbb{R}^{+}, \nu \in \left(0,1\right], \lambda\in \mathbb{R}^+_0$}

\REPEAT
	\STATE{$t \leftarrow t + 1$}
	\smallskip
	
	\STATE{$\tilde{\bm{\theta}} \leftarrow \bm{\theta}_{t-1} + \alpha\bm{v}^w_{t-1} $}
	\STATE{$\tilde{\bm{b}} \leftarrow \bm{b}_{t-1} + \alpha\bm{v}^b_{t-1} $}
	\medskip
	\STATE{$\bm{g}^{\bm{\theta}}_t \leftarrow \frac{1}{m} \nabla_{\tilde{\bm{\theta}}} \sum_i L\left(f(x^{(i)};\tilde{\bm{\theta}}),y^{(i)}\right)$  \hfill  \text{ kernel gradients}  }
	
	\STATE{$\bm{g}^b_t \leftarrow \frac{1}{m}\nabla_{\tilde{\bm{b}}} \sum_i L\left(f(x^{(i)};\tilde{\bm{b}}),y^{(i)}\right)$  \hfill  \text{ bias gradients} }
	\medskip
	
	\STATE{$\bm{v^{\bm{\theta}}_t} \leftarrow \rho\bm{v^{\bm{\theta}}_t} +  h\left(\bm{g}^{\bm{\theta}}_t\right)$}
	
	\STATE{$\bm{v^{b}_t} \leftarrow \rho\bm{v^{b}_t}+  \bm{g}^w_t$}
	\medskip
	
	\STATE{$\bm{\theta}_t \leftarrow \bm{\theta}_{t-1} - \alpha\bm{v^{w}_t}   - \lambda\bm{\theta}_{t-1} $} 
	
	\STATE{$\bm{\theta}_t \leftarrow \bm{b}_{t-1} - \alpha\bm{v^{b}_t}   - \lambda\bm{b}_{t-1} $} 
	
	\medskip
\UNTIL{ \textit{stopping criterion is met}}
\end{algorithmic}
\end{algorithm}

\subsection{Behaviour near a minimum}

It should be clear that as we get closer to the minimum, and the gradient becomes shallower, the amplifying effect of the nonlinear function will increase. It is likely that applying a NL function will see noisier behaviour at convergence than, for example, SGD. This could be a weakness of the approach; however, we know that there are already other mechanisms that produce similar effects. For example, the stochastic behaviour with small batch sizes or overly large learning rates can produce overshooting when close to convergence. This is demonstrated in the benefit of decaying learning rates or increasing of batch size \cite{increase_batch_size}. So, at worst, we are exacerbating the problem rather than introducing something new. \par 

To account for this behaviour near the minima, we can take an approach similar to that in \cite{superconvergence} and reduce the learning rate significantly, close to the end of training -- frequently referred to as an `annihilation' phase. Generally, we find that such a phase does improve performance with NL algorithms; in the tests below we will identify some cases where this addition is particularly notable.\par

\section{Benchmarking Overview}

Our primary tool for assessing the performance of optimizers with fractional exponents is the DeepOBS benchmark \cite{deepobs}. This benchmark provides infrastrucure and a set of reference cases to measure optimizer performance. A subset of the reference cases were used in \cite{crowdedvalley} to assess the performance of a large set of commonly used optimizers. Consequently this provides us with a clear comparison point. \par

The test cases used in \cite{crowdedvalley} are
\begin{enumerate}
    \item[\textbf{P1}] Quadratic Deep
    \item[\textbf{P2}] VAE applied to MNIST 
    \item[\textbf{P3}] Simple CNN applied to F-MNIST
    \item[\textbf{P4}] Simple CNN applied to CIFAR-10
    \item[\textbf{P5}] VAE applied to F-MNIST 
    \item[\textbf{P6}] CNN applied to CIFAR-100
    \item[\textbf{P7}] Wide ResNet appplied to SVHN
    \item[\textbf{P8}] RNN applied to War and Peace
\end{enumerate}

These test cases are partitioned into `small' (P1 to P4) and `large' (P5 to P8). The small examples allow rapid investigations - and we will use them to identify some characteristics of the \nl{} optimizer.\par

To allow exploration of the impact of different mechanisms, $L_2$ regularisation/weight decay, dropout \cite{Srivastava_dropout} and batch normalisation \cite{Nitish_Shirish_Keskar_et_al_2016} are applied in some cases and not in others. See text below for details in each case.\par

In addition, \cite{crowdedvalley} aims to replicate the process of hyperparameter section with a selection of randomised search algorithms applied to all optimizers. 
Three algorithms with increasing search depth were considered - `small', `medium' and `large'. For the investigation here, we focus on the `medium' option. In practice, we found that this gives a good indication of optimizer performance without excessive search time.\par

Medium randomisation takes 50 samples of the hyperparameters for each optimizer and then generates one training run for each same. A validation set is used to select the best hyperparameter set. For the purpose of evaluating the optimzer, the hyperparameter sample with the best validation accuracy or loss (depending on the test case) is selected and training runs with this setting are repeated 10 times, with different random seeds.   \par

The distributions were as follows:

\begin{tabular}{ l  l  l}
	
		 \textbf{Parameter}  && \textbf{Tuning Distribution} \\ 
		 learning rate &$\alpha$&\text{log-uniform distribution}\\
		 &&$\mathcal{LU}(10^{-4}, 1)$  
		 \smallskip{}\\
		 momentum&$\rho$&\text{fixed at $0.9$}\\
         exponent&$\nu$&\text{selected uniformly from}\\
		 &&$\left\{0.4, 0.5, 0.6, 0.7, 0.8, 0.9\right\}$\\

\end{tabular}

Of course, the search could be expanded beyond this, most obviously by including additional options for the momentum parameter and a continuous range for the exponent. For this initial investigation, the decision was to avoid the search space becoming too large. \par

Ref \cite{crowdedvalley} also considered a number of different learning rate schedules. For the purpose of this paper, we concentrate primarily on fixed learning rates. However, as already noted, where informative, we will provide results with an `annihilation' phase; we implement this as a reduction in the learning rate by 10x for the last 5 epochs of training. (This is predominantly to show the benefit of counteracting the `noise amplification' close to convergence rather than to imply this is of unique benefit for NL algorithms.)\par

For each test case, we extract five results from  \cite{crowdedvalley} for comparison: SGD, Momentum, NAG, Adam, plus the best performing result for the test case being considered.\par
For the \nl{} results we give three variants, corresponding to the algorithms outlined above: \nl{}, \nlm{} and \nln{}.\par

Note that, for assessing performance in a consistent way with the results of \cite{crowdedvalley}, we want the weight decay term in our NL training to have a comparable magnitude to the $L_2$ term used previously. For \nl{} this is relatively straightforward, since we can simply set $\lambda = \alpha\lambda'$, where $\lambda'$ is the value reported in \cite{crowdedvalley}. For \nlm{} and \nln{}, we must recognise that the weight decay term will have an amplification due to the momentum mechanism. This complicates selecting a value for $\lambda$ for \nlm{} and \nln{} as it is not possible to pick an exactly corresponding value. For the comparisons here, we use $\lambda = \alpha\lambda'/\left(1-\rho\right)$. This approximates the momentum amplification as an infinite geometric series in the $\rho$ term, which would be correct if the weight magnitude did not change. Since we restrict $\rho$ to 0.9 for \nlm{} and \nln{} in this discussion, the momentum contribution drops quickly and the comparison is considered reasonable.\par

We note at the outset that \snr{} was constructed with image classification as the primary focus. These have clear features consisting of correlated inputs and will be affected by noise. Consequently, we expect the four classification test cases to be those where \nl{} brings the most benefit. However, it is instructive to see whether there is wider applicability.\par
 
All test were carried out using Tensorflow \cite{tensorflow2015-whitepaper}.

\section{DeepObs -- Small Test Cases}

In this section, we consider the four small test problems. The ability to train these models quickly allows us to do a more thorough investigation into the relationship between performance and hyperparameter dependency. \par 

For each test case we give a very brief description identifying key points on the construction and regularisation configuration. For further details, the reader should consult \cite{deepobs}. 

\subsection{F-MNIST 2C2D}

FMNIST 2C2D is a CNN, with 2 convolutional layers and two fully connected layers applied to the Fashion MNIST dataset \cite{fmnist}. No weight decay, dropout or batch normalisation is applied. The training is carried out for 100 epochs with a batch size of 128. \par

The results are given in table \ref{tab:fmnist2c2d}. The immediate observation is that each NL algorithm performs better then the standard equivalent. \nln{} comes close to the best performing optimizer in \cite{crowdedvalley}. \par

{
    \centering
    \begin{threeparttable}
    \begin{tabular}[t]{lccc}
    \hline
    &Test Accuracy & \demph{$\nu$}&\demph{$1-\rho$}\\
    \hline
    SGD&      $91.76 \pm 0.19$ & \demph{-}& \demph{-}\\
    Momentum& $91.91 \pm 0.14$ &  \demph{-}& \demph{0.01}\\
    NAG&      $92.13 \pm 0.14$ &  \demph{-}& \demph{0.01}\\
    Adam&     $91.61 \pm 0.28$ & \demph{-}& \demph{-}\\
    Best (AMSGrad)& $92.27 \pm 0.18$ & \demph{-}& \demph{-}\\
    \hline
    \nl{}&    $92.04 \pm 0.20$ &\demph{0.5}& \demph{-}\\
    \nlm{}&   $92.16 \pm 0.16$ &\demph{0.7}& \demph{0.1}\\
    \nln{}&   $92.20 \pm 0.17$ &\demph{0.7}& \demph{0.1}\\
    \hline
    \end{tabular}
    \caption{F-MNIST 2C2D}
    \label{tab:fmnist2c2d}

    \end{threeparttable}
    \par
}

\begin{figure}[!htbp]
    \centering
    \begin{subfigure}[b]{0.95\linewidth}
        \includegraphics[width=\linewidth]{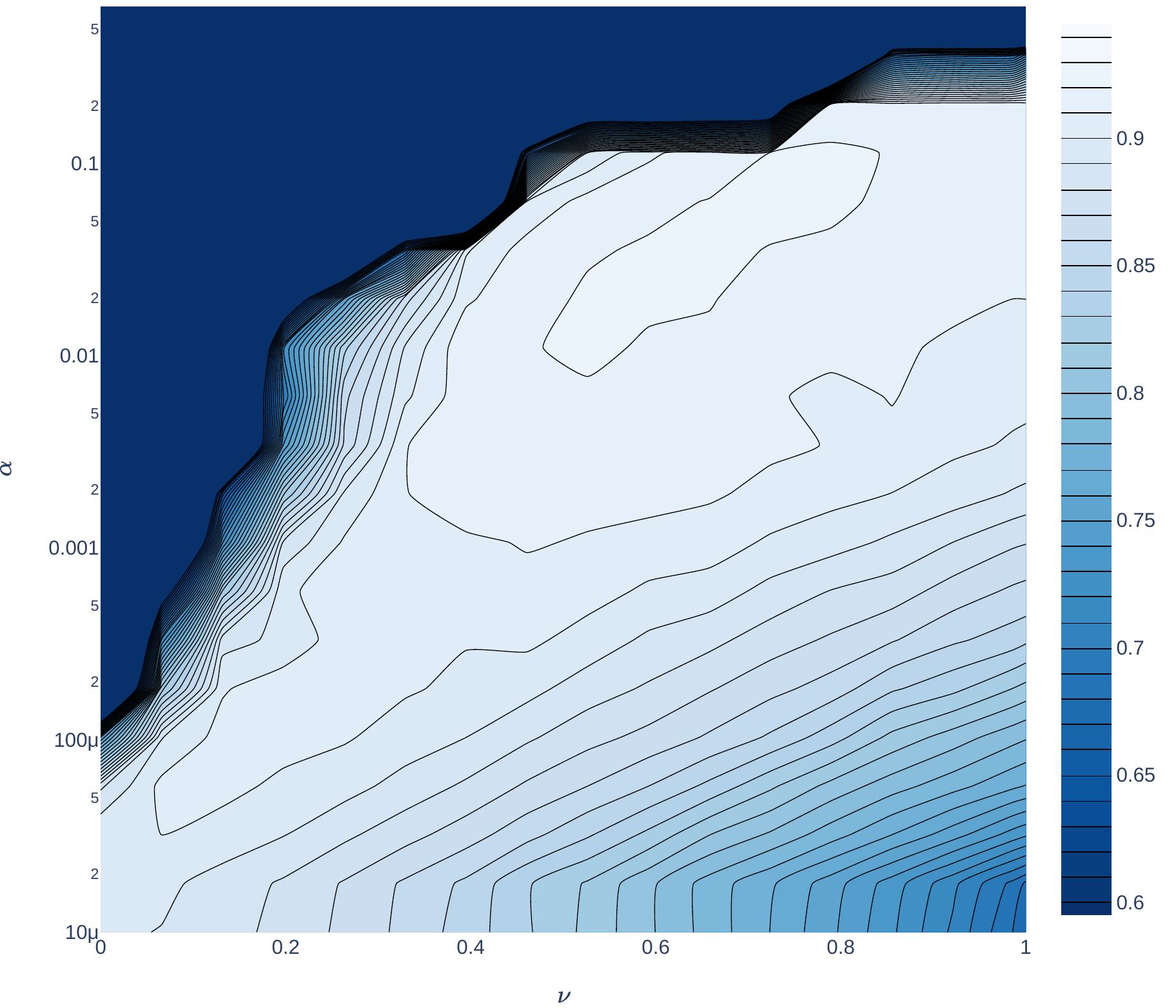}
        \caption{SGD}

    \end{subfigure}
     
    \par
     
    \begin{subfigure}[b]{0.95\linewidth}
       \includegraphics[width=\linewidth]{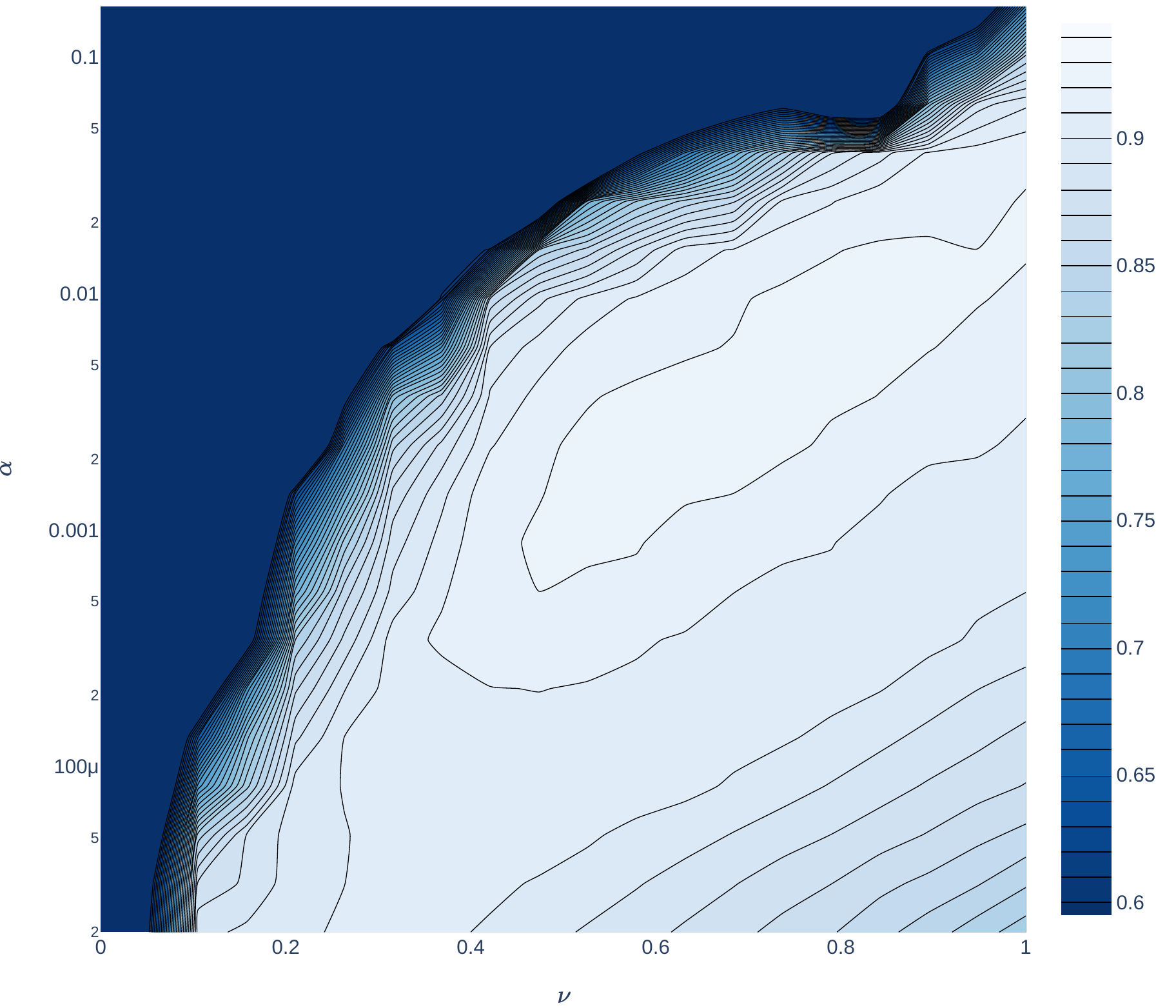}
       \caption{Momentum 0.9}
    \end{subfigure}

    \caption{Hyperparameter dependancies for FMNIST 2c2d}
    \label{fig:fmnist2c2d}
\end{figure}

It is instructive to plot the performance errors (validation set) for different combinations of $\nu$ and $\alpha$, see figure \ref{fig:fmnist2c2d}. This demonstrates that the optimal meta-parameter set is one with $\nu<1$ and, in fact, any value for $\nu$ with $\nu\in [0.5, 1)$ will allow some improvement over conventional SGD or Momentum\footnote{The plot also gives confirmation that, for this simple case, with very similar performance between optimizers, the improvement in the NL versions is a real effect and not simply statistical variation.}.\par

It is also notable that there is approximately a linear relationship between $\nu$ and log of the peak learning rate at that exponent.\par

\subsection{CIFAR-10 3C3D}

This test case applies a CNN with 3 convolution layers and 3 dense layers to the CIFAR-10 dataset. The activation functions are ReLU and the training uses $L_2$ regularisation (updated to weight decay for the NL optimzers).\cite{CIFAR}. The training is carried out for 100 epochs with a batch size of 128.\par

{
    \centering
    \begin{threeparttable}
    \begin{tabular}[t]{lccc}
    \hline
    &Test Accuracy &\demph{$\nu$}&\demph{$1-\rho$}\\
    \hline
     SGD&      $82.37 \pm 1.16$ & \demph{-}&\demph{-}\\
     Momentum& $83.00 \pm 0.86$ & \demph{-}&\demph{0.01}\\
     NAG& $83.43 \pm 0.55$ & \demph{-}&\demph{0.14}\\
     Adam&     $82.82 \pm 0.91$ & \demph{-}&\demph{-}\\
     Best (NAG)&  $83.43 \pm 0.55$&  \demph{-}&\demph{0.14}\\
     \hline
     \nl{}&  $83.24 \pm 0.79$&\demph{0.7}&\demph{-}\\
     \nlm{}& $83.83 \pm 0.35$&\demph{0.7}&\demph{0.9}\\
     \nln{}& $84.21 \pm 0.53$&\demph{0.7}&\demph{0.9}\\
    \hline
    \end{tabular}
    \caption{CIFAR-10 3C3D}
    \label{tab:cifar3c3d}

    \end{threeparttable}
    \par
}

The results are given in table \ref{tab:cifar3c3d}. Again we observe that each NL algorithm performs better then the standard equivalent. In this case, both \nlm{} and \nln{} perform better than all other optimizers.\par

We should be careful to rule out the possibility this improvement has come, not from the use of the NL function, but purely from applying weight decay to Momentum instead of $L_2$ regularisation. With this in mind, we note that even if we revert to the $L_2$ approach, the mean accuracy for \nlm{} is $83.55$, which still outperforms other optimizers.\par

Again, plotting performance with varying $\nu$ and $\alpha$ gives comparable results to the previous test case, most markedly for \nlm{}.\par

\begin{figure}[!htbp]
\centering
  
     \begin{subfigure}[b]{0.95\linewidth}
       \includegraphics[width=\linewidth]{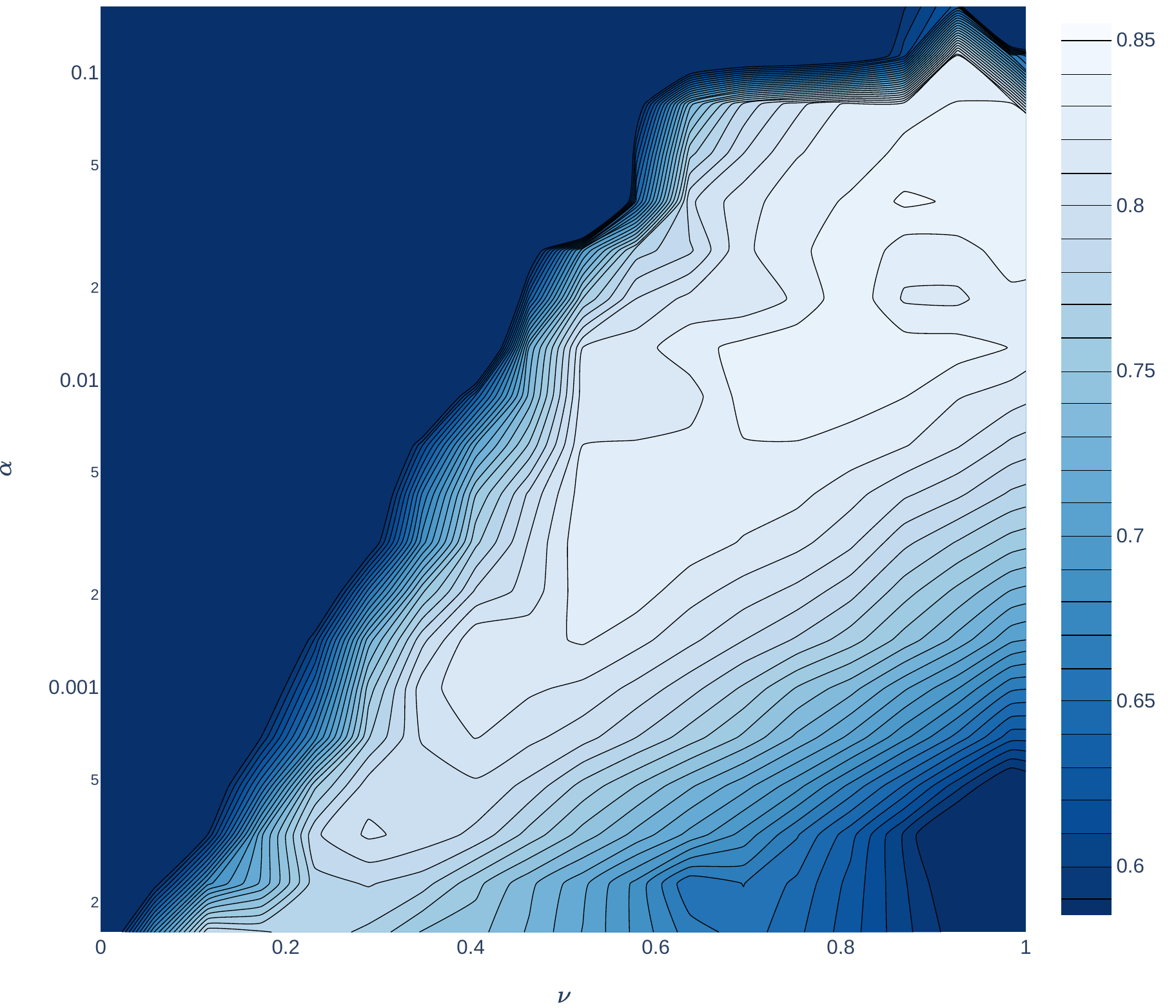}
       \caption{SGD}

     \end{subfigure}
     \par
     \begin{subfigure}[b]{0.95\linewidth}
       \includegraphics[width=\linewidth]{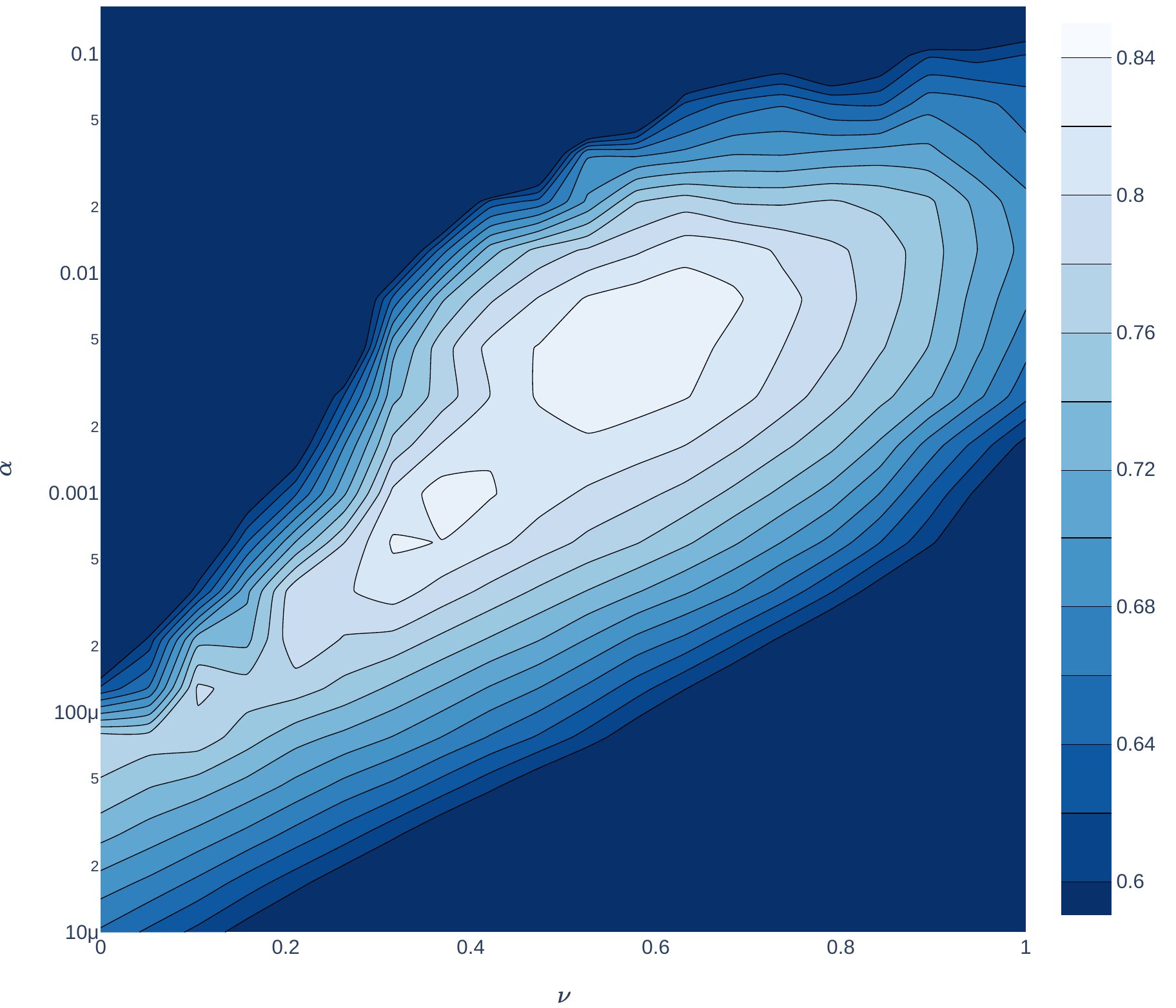}
       \caption{Momentum 0.9}
     \end{subfigure}
     
\caption{Hyperparameter dependancies for CIFAR-10 3c3d}
     \label{fig:cifar10}
   \end{figure}

If we introduce an `annihilation' phase at the end of training then we see some further improvement\footnote{Note that this was not obtained with an addition hyperparameter search, but simply took the best result from the previous search and reran the test cases with the added scheduling.}: \nl{}, \nlm{} and \nln{} achieve test accuracies of $84.21$, $85.56$ and $85.75$.

\subsection{MNIST VAE}
\label{sec:mnist_vae}
This test case applies a VAE to the MNIST dataset \cite{MNIST, lecun2010mnist}. The convolutional layers use a leaky ReLU activation function and dropout is applied after each convolutional layer. The training is carried out for 50 epochs with a batch size of 64.\par

In this case, the NL algorithms have more stable results than the non-NL variants, table \ref{tab:mnistvae}. (SGD and Momentum all have large variance, reflecting the fact that they sometimes fail to find a good minimum with the selected hyperparameters.) However, none of the NL variants achieve performance close to Adam. \par

{    \centering
    
    \begin{threeparttable}
   
    \begin{tabular}[t]{lccc}
    
    \hline
    &Test Loss&\demph{$\nu$}&\demph{$1-\rho$}\\
    \hline
    SGD&      $\phantom{0}36.18 \pm 10.96$& \demph{-}& \demph{-} \\
    Momentum& $\phantom{0}35.99 \pm 11.09$& & \demph{0.64}  \\
    NAG&      $\phantom{0}36.23 \pm 10.93$& & \demph{0.64}  \\
    Adam&     $27.77 \pm  0.07$& \demph{-}& \demph{-} \\
    Best (Adam)& $27.77 \pm 0.07$& \demph{-}& \demph{-} \\
     \hline
     \nl{}& $28.41 \pm 0.13$& \demph{0.6}& \demph{-} \\
    \nlm{}& $28.57 \pm 0.12$& \demph{0.5}& \demph{0.1} \\
    \nln{}& $28.46 \pm 0.19$& \demph{0.5}& \demph{0.1} \\
    \hline
    \end{tabular}
    \caption{MNIST VAE}
    \label{tab:mnistvae}
    \begin{tablenotes}\footnotesize
    \item{Note: SGD, Momentum and NAG all see high losses and variance due to a mix with some runs that converge well and some that converge badly.}
    \end{tablenotes}
    \end{threeparttable}
    \par
}

\begin{figure}[!htbp]
      \centering
     \begin{subfigure}[b]{0.95\linewidth}
       \includegraphics[width=\linewidth]{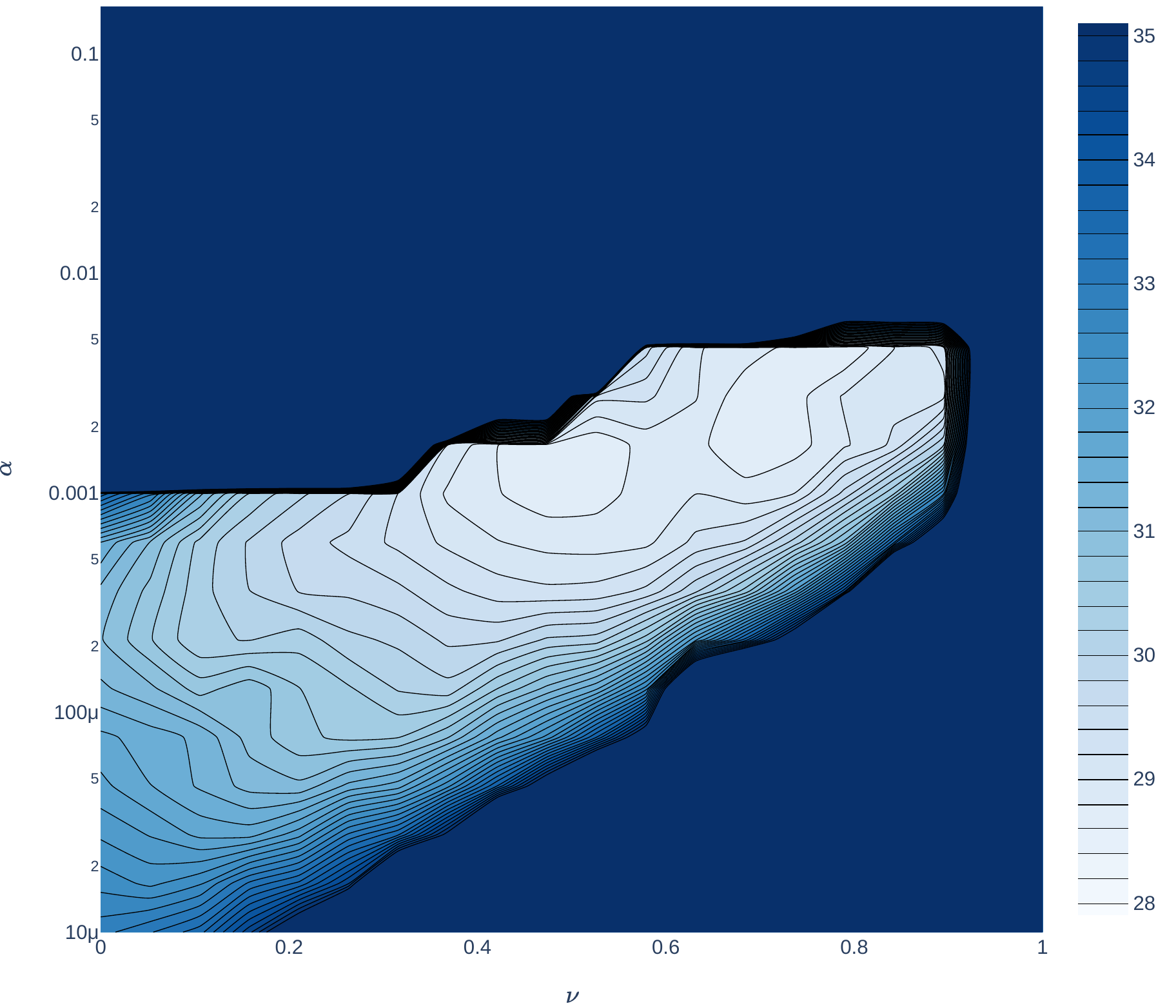}
       \caption{SGD}

     \end{subfigure}
    \par
     \begin{subfigure}[b]{0.95\linewidth}
       \includegraphics[width=\linewidth]{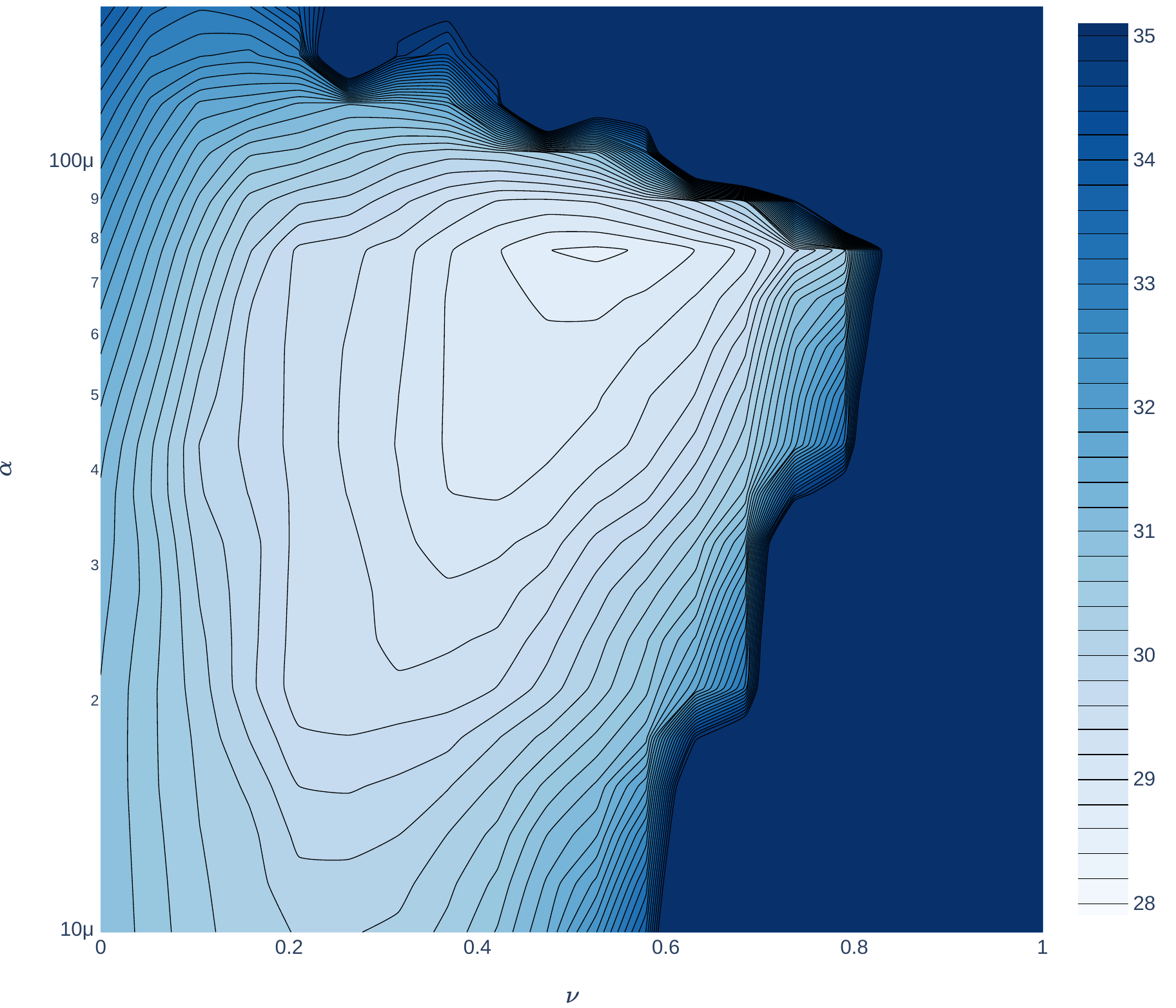}
       \caption{Momentum 0.9}
     \end{subfigure}

\caption{Hyperparameter dependancies for MNIST VAE}
     \label{fig:MNIST_simple}
   \end{figure}

As with previous test cases, we can look at the performance for different $\nu-\alpha$ combinations. The results for \nl{} and \nlm{} are shown in figure \ref{fig:MNIST_simple}. This suggests that the better performance of NL algorithms compared to SGD, Momentum and NAG is not simply due to the random search finding better points, but that applying the NL function allows a more stable training region to be found. \par

With an `annihilation' phase at the end of training \nl{} and \nlm{} achieve test accuracies of $28.08$ and $28.33$.

\subsection{Quadratic Deep}
The quadratic deep test problem generates the test data from as a vector of IDD normally distributed samples and a shallow function. 

\begin{equation}
   \Loss =  \frac{1}{2} (\theta - x)^T Q (\theta - x)
\end{equation}
 with Hessian Q, trainable parameters $\theta$ and "data" $x$ a vector if IDD samples from a zero-mean normal.\par
 
 $Q$ is constructed such that 90\% of the eigenvalues of the Hessian are drawn from the interval $\left(0.0, 1.0\right)$ and the other 10\% are from $\left(30.0, 60.0\right)$. The intention is to construct an eigenspectrum comparable with that discussed in \cite{biaswidevalleys}. \par
 The training is carried out for 100 `epochs' with a batch size of 128.\par

Considering that our initial motivation came from the presence of correlated inputs into the network or into the hidden layers, we would not expect application of a NL function to improve the training for this model. This is exactly what we observe. The results in table \ref{tab:quadraticdeep} show that none of the NL variants provide an advantage. In fact, they perform marginally worse than most of the other optimizers assessed in \cite{deepobs}.\par

{
    \centering
    \begin{threeparttable}
    \begin{tabular}[t]{lccc}
    \hline
    &Test Loss &\demph{$\nu$}&\demph{$1-\rho$}\\
    \hline
    SGD&      $86.29 \pm 3.44$& \demph{-}& \demph{-} \\
    Momentum& $87.02 \pm 0.02$&  \demph{-}& \demph{0.01} \\
    NAG &    $87.08 \pm 0.02$&  \demph{-}& \demph{0.01} \\
     Adam&    $86.58 \pm 1.95$& \demph{-}& \demph{-} \\
     Best (SGD)&  $86.29 \pm 3.44$& \demph{-}& \demph{-} \\
     \hline
    \nl{}&    $87.69 \pm 0.10$& \demph{0.8}& \demph{-} \\
    \nlm{}&   $87.14 \pm 0.06$& \demph{0.9}& \demph{0.9} \\
    \nln{}&   $87.17 \pm 0.06$& \demph{0.9}& \demph{0.9} \\
    \hline
    \end{tabular}
    \caption{Quadratic Deep}
 \label{tab:quadraticdeep}
    \end{threeparttable}
    \par
}

\begin{figure}[!htbp]
      \centering
     \begin{subfigure}[b]{0.95\linewidth}
       \includegraphics[width=\linewidth]{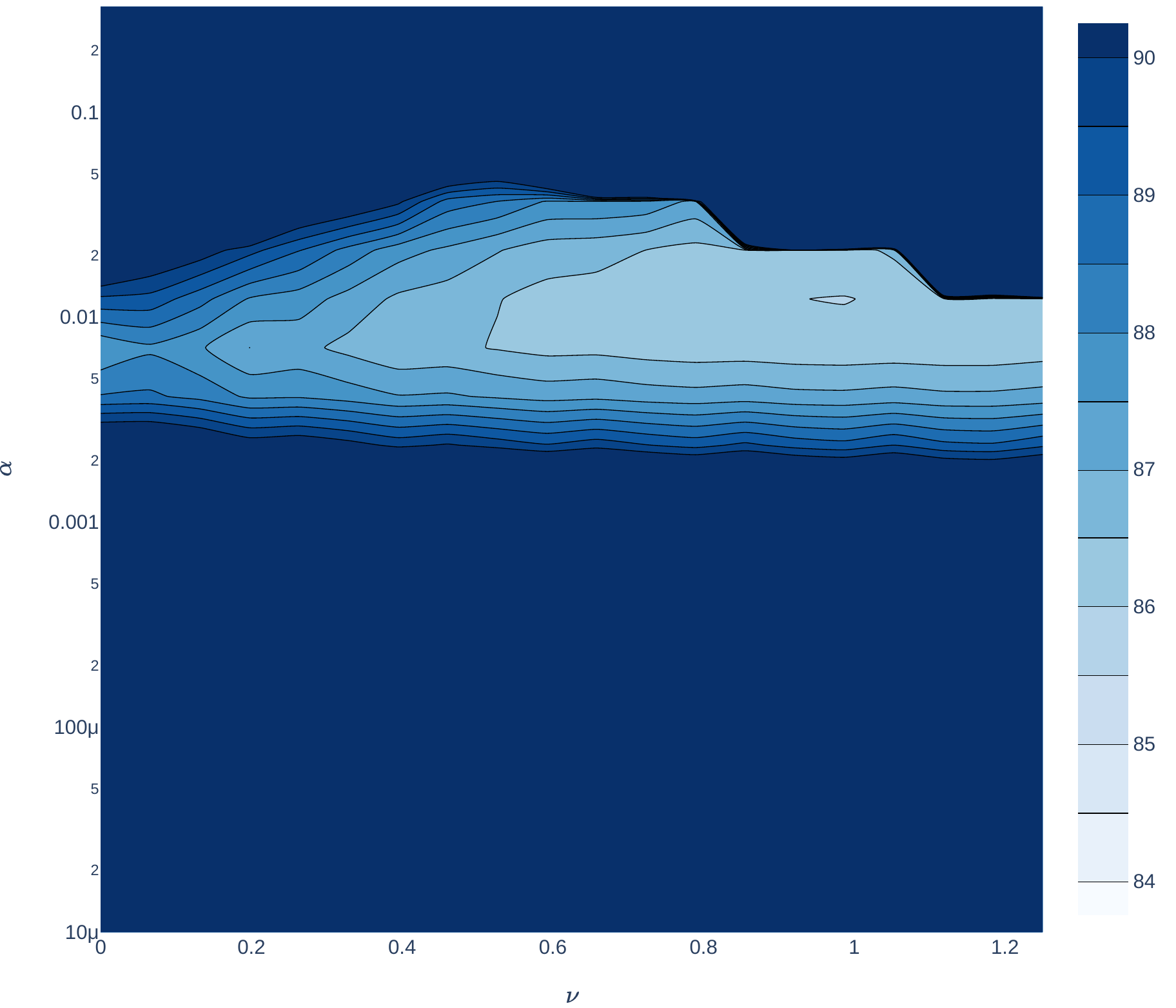}
       \caption{SGD}

     \end{subfigure}
    \par
     \begin{subfigure}[b]{0.95\linewidth}
       \includegraphics[width=\linewidth]{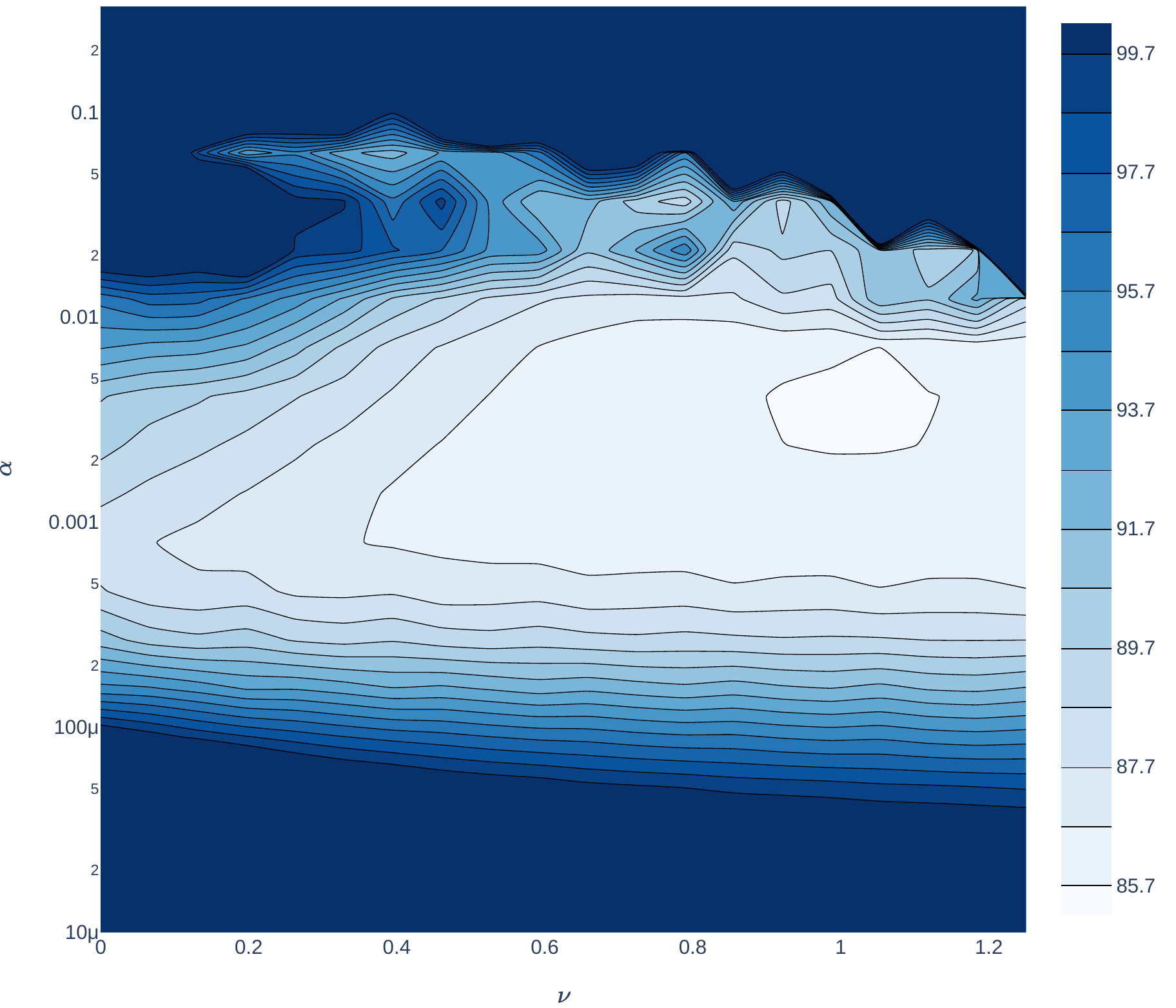}
       \caption{Momentum 0.9}
     \end{subfigure}

\caption{Hyperparameter dependencies for Quadratic Deep}
     \label{fig:quadratic_deep}
   \end{figure}

Figure \ref{fig:quadratic_deep} shows the performance as the two hyperparameters are varied. Consistent with expectations, the optimal points correspond to $\nu = 1$ -- that is, basic SGD and Momentum -- with performance dropping off as the exponent is reduced (or, indeed, increased). Further, the optimal region is relatively small, which suggests why the random search process was unable to find a good configuration. \par

If we include an annihilation phase then \nl{} and \nlm{} reach losses of $86.29$ and $85.71$, respectively. For this simple case, the implication is that the optimizers are overshooting due to amplification of stochastic effects.\par

\subsection{Discussion}
In every test case, apart from Quadratic Deep, we see that applying the NL function in conjunction with SGD, Momentum or NAG leads to better performance that the  usual implementation of the optimizers. In fact, the sensitivity plots indicate that even a small non-linearity will improve performance. Obviously, this could be applied without the search over $\nu$ considered here. \par

The exception effectively proves the rule, since Quadratic Deep does not fit the criteria that inspired the NL approach. \par

We note that, in most cases considered, the relationship between the exponent hyperparameters and optimal learning rate for that exponent is simple. Although not addressed in detail here, this suggests that an efficient search approach could be defined to find a good combination. \par 

\newpage
\section{DeepObs -- Large Test Cases}

In this section we consider results from the DeepObs `large' test cases. \par

\subsection{CIFAR-100 All CNN C}

This test case applies a variation of the `All Convolutional Net' \cite{allcnn} to the CIFAR-100 dataset \cite{CIFAR}. The activation functions are ReLU and the training configuration includes dropout. The training is carried out for 350 epochs with a batch size of 256.\par

As in previous cases, we see that the NL versions of optimizers out-perform their non-NL counterparts, see table \ref{tab:cifar100acnnc}. In this case, \nlm{} performs better than any of the optimizers considered in \cite{crowdedvalley}.\par

{
    \centering
    \begin{threeparttable}
    \begin{tabular}[t]{lcccc}
    \hline
    &Test Accuracy&\demph{$\nu$}&\demph{$1-\rho$}\\
    \hline
    SGD&      $50.71 \pm 1.85$&\demph{-}& \demph{-} \\
    Momentum& $55.92 \pm 0.82$& \demph{-}& \demph{3e-4}\\
    NAG&      $56.94 \pm 0.73$& \demph{-}& \demph{3e-4}\\
    Adam&     $53.93 \pm 0.69$&\demph{-}& \demph{-} \\
    Best (LA Mom)& $57.49 \pm 1.16$& \demph{-}\\
    \hline
    \nl{}&    $55.05 \pm 0.92$& \demph{0.6}& \demph{-} \\
    \nlm{}&   $58.31 \pm 0.80$& \demph{0.9}& \demph{0.1} \\
    \nln{}&   $57.43 \pm 0.64$& \demph{0.8}& \demph{0.1} \\
    \hline
    \end{tabular}

       \caption{CIFAR-100 All CNN C}
       \label{tab:cifar100acnnc}
    \end{threeparttable}
    \par
}
\subsection{SVHN WRN}

This test case implements a Wide ResNet 16-4 architecture, as proposed in \cite{wrn} and applied to the SVHN dataset \cite{svhn}. The activation functions are ReLU and the network includes batch normalisation. The training is carried out for 160 epochs with a batch size of 128.\par

In this case, all the NL algorithms perform better than all of the optimizers considered in \cite{crowdedvalley}, see table \ref{tab:svhnwrn}.\par

{
    \centering
    \begin{threeparttable}
    \begin{tabular}[t]{lcccc}
    \hline
    &Test Accuracy&\demph{$\nu$}&\demph{$1-\rho$}\\
    \hline
    SGD&      $94.88 \pm 0.67$ & \demph{-}&\demph{-}\\
    Momentum& $95.30 \pm 0.19$ & \demph{-}&\demph{0.01} \\
    NAG&      $95.45 \pm 0.23$ & \demph{-} &\demph{0.01}\\
    Adam&     $95.16 \pm 0.32$ & \demph{-}&\demph{-}\\
    Best (AMS Bound)& $95.45 \pm 0.14$ &\demph{-}\\
    \hline
    \nl{}& $96.30 \pm 0.15$& \demph{0.5}&\demph{-}\\
    \nlm{}& $96.29 \pm 0.07$& \demph{0.4}&\demph{0.1}\\
    \nln{}& $96.40 \pm 0.08$& \demph{0.4}&\demph{0.1}\\
    \hline
    \end{tabular}

    \caption{SVHN WRN}
    \label{tab:svhnwrn}
    \end{threeparttable}
    \par
}

\subsection{F-MNIST VAE}

In this case applies, the VAE used above in section \ref{sec:mnist_vae} is applied to the Fashion MNIST dataset. The training is carried out for 100 epochs with a batch size of 64.\par

As for the MNIST version, the results are better than the non-NL versions of the algorithms, but do not match the performance of Adam-like optimizers, see table \ref{tab:fmnistvae}. \par

{
    \centering
    \begin{threeparttable}
    \begin{tabular}[t]{lccc}
    \hline
    &Test Loss&\demph{$\nu$}&\demph{$1-\rho$}\\
    \hline
    SGD&      $23.88 \pm 0.31$&\demph{-}\\
    Momentum&  $\phantom{0}28.36 \pm 13.20$&\demph{-}& \demph{0.07}  \\
    NAG&      $23.89 \pm 0.14$&\demph{-}& \demph{0.07}  \\
    Adam&     $23.06 \pm 0.09$&\demph{-}& \demph{-}\\
    Best (NADAM)& $23.01 \pm 0.07$&\demph{-}& \demph{-}\\
    \hline
    \nl{}&  $23.71 \pm 0.14$& \demph{0.9}&\demph{-}\\
    \nlm{}& $\phantom{0}28.19 \pm 13.47$& \demph{0.8}& \demph{0.1}\\
    \nln{}& $23.56 \pm 0.18$& \demph{0.5}& \demph{0.1}\\
    \hline
    \end{tabular}
    \begin{tablenotes}\footnotesize
    \item{Note: Momentum and \nlm{} results are both affected by a single case that failed to converge. Momentum found a better solution of $23.86 \pm 0.15$ in the large search.}
    \end{tablenotes}
        \caption{F-MNIST VAE}
\label{tab:fmnistvae}
    \end{threeparttable}
    
    \par
}

\subsection{Tolstoi (War and Peace RNN)}

The final test case applies an RNN for character prediction on War and Peace. It is build from the standard Tensorflow LSTMCell and MultiRNNCell, with tanh activation functions and dropout. The training is carried out for 200 epochs with a batch size of 50.\par

{
    \centering
    \begin{threeparttable}
    \begin{tabular}[t]{lccc}
    \hline
    &Test Accuracy&\demph{$\nu$}&\demph{$1-\rho$}\\
    \hline
    SGD&      $61.23 \pm 0.12$ & \demph{-} & \demph{-}\\
    Momentum& $61.91 \pm 0.11$ & \demph{-}& \demph{0.17}\\
    NAG&      $61.93 \pm 0.11$ & \demph{-}&\demph{0.13}\\
    Adam&     $61.96 \pm 0.11$ & \demph{-}& \demph{-}\\
    Best (RMSProp)& $62.24 \pm 0.10$& \demph{-}& \demph{-}\\
    \hline
    \nl{}&   $61.86 \pm 0.08$& \demph{0.7}& \demph{-}\\
    \nlm{}&   $61.39 \pm 0.13$& \demph{0.6}& \demph{0.1}\\
    \nln{}&   $61.57 \pm 0.07$& \demph{0.6}& \demph{0.1}\\
    \hline
    \end{tabular}

    \caption{Tolstoi}
    \label{tab:tolstoi}
    \end{threeparttable}
    \par
}

This is the only case, in addition to Quadratic Deep, where the \nlm{} and \nln{} algorithms perform worse than their conventional equivalents (table \ref{tab:tolstoi}). It is beyond the scope of this discussion to identify the reasons for this behaviour, but results suggest that the combination of the NL function, higher momentum and dropout add instabilities that impede good performance. Compared to the linear equivalents, both algorithms show a higher level of instability during training when the training rate is large. This makes it harder to find an optimal point by random parameter selection. It is possible to improve performance by tuning the parameters by hand (following an approach similar to \cite{smith_cyclic}, we can reach an accuracy of 61.95 with \nln{}) or removing dropout and re-searching (\nln{} gives an accuracy of 61.97), but neither give the improvement seen in other cases. \par

\section{Conclusion}

We have presented an adaptation for DNN optimizers, which applies a non-linear function to gradients prior to applying them to the weights of a network. This was inspired by the work in \cite{snr} which showed that robustness to noise can be a significant contributor to generalisation performance. Our non-linear approach aims to encourage enhanced SNR-optimal performance by balancing the weights for correlated inputs to a node.\par

The results show that, with a few exceptions, the application of a NL function to weight updates does generate better performance than with the same optimizer and no NL application. For the larger image classification tasks, NL algorithms give better results than all other optimizers characterised in \cite{crowdedvalley}. A benefit of this approach is that  it retains a small memory footprint compared to, for example, Adam variants.   \par

Of course, there are many possible extensions of the work presented here: we have by no means explored all the NL functions that could provide benefit, and the application of the NL adaptation could be applied to other optimizers. \par 

The intention of this investigation was to extend the study in \cite{snr}. We have seen that the SNR-inspired approach improves performance for exactly the cases where it is most relevant, image classification, and provides no benefit in the test case where the background assumptions (correlated inputs to nodes) are not met. While this does not demonstrate the validity of \snr{} unequivocally, it does suggest that it is a productive way of looking at optimization.\par 

\paragraph{Acknowledgments}

Thanks to Frank Schneider for clarifications on the DeepObs benchmarking and to Max Pudney for reading the draft.\par

Full results will be available at \url{http://github.com/pnorridge/nonlinear-weight-updates}

\printbibliography

@ARTICLE{2017arXiv171105101L,
       author = {{Loshchilov}, Ilya and {Hutter}, Frank},
        title = "{Decoupled Weight Decay Regularization}",
      journal = {arXiv e-prints},
         year = 2017,
        month = nov,
          eid = {arXiv:1711.05101},
        pages = {arXiv:1711.05101},
archivePrefix = {arXiv},
       eprint = {1711.05101},
 primaryClass = {cs.LG},
       adsurl = {https://ui.adsabs.harvard.edu/abs/2017arXiv171105101L}
}

@article{Chiyuan_Zhang_et_al_2016,

       author = {{Zhang}, Chiyuan and {Bengio}, Samy and {Hardt}, Moritz and
         {Recht}, Benjamin and {Vinyals}, Oriol},
        title = "{Understanding deep learning requires rethinking generalization}",
      journal = {arXiv e-prints},
         year = "2016",
        month = "Nov",
          eid = {arXiv:1611.03530},
archivePrefix = {arXiv},
       eprint = {1611.03530},
 primaryClass = {cs.LG},
       adsurl = {https://ui.adsabs.harvard.edu/abs/2016arXiv161103530Z}
}

@article{snr, 
  author    = {Paul Norridge},
  title     = {Think Global, Act Local: Relating {DNN} generalisation and node-level
               {SNR}},
  journal   = {CoRR},
  volume    = {abs/2002.04687},
  year      = {2020},
  url       = {https://arxiv.org/abs/2002.04687},
  eprinttype = {arXiv},
  eprint    = {2002.04687},
  bibsource = {dblp computer science bibliography, https://dblp.org}
}

@article{Sepp_Hochreiter_Jrgen_Schmidhuber1997,
 author  = {Sepp Hochreiter and
      J{\"u}rgen Schmidhuber},
 title  = {Flat Minima},
 journaltitle = {Neural Computation},
 year  = {1997},
 month  = {1},
 publisher = {MIT Press - Journals},
 volume  = {9},
 number  = {1},
 doi  = {10.1162/neco.1997.9.1.1},
 url  = {http://dx.doi.org/10.1162/neco.1997.9.1.1},
 pages  = {1-42},
 issn  = {0899-7667},
}

@article{Nitish_Shirish_Keskar_et_al_2016,
       author = {{Shirish Keskar}, Nitish and {Mudigere}, Dheevatsa and {Nocedal}, Jorge and
         {Smelyanskiy}, Mikhail and {Tang}, Ping Tak Peter},
        title = "{On Large-Batch Training for Deep Learning: Generalization Gap and Sharp Minima}",
      journal = {arXiv e-prints},
         year = "2016",
        month = "Sep",
          eid = {arXiv:1609.04836},
archivePrefix = {arXiv},
       eprint = {1609.04836},
 primaryClass = {cs.LG},
       adsurl = {https://ui.adsabs.harvard.edu/abs/2016arXiv160904836S}
}

@misc{tensorflow2015-whitepaper,
title={ {TensorFlow}: Large-Scale Machine Learning on Heterogeneous Systems},
url={https://www.tensorflow.org/},
note={Software available from tensorflow.org},
author={
    Mart\'{i}n~Abadi and
    Ashish~Agarwal and
    Paul~Barham and
    Eugene~Brevdo and
    Zhifeng~Chen and
    Craig~Citro and
    Greg~S.~Corrado and
    Andy~Davis and
    Jeffrey~Dean and
    Matthieu~Devin and
    Sanjay~Ghemawat and
    Ian~Goodfellow and
    Andrew~Harp and
    Geoffrey~Irving and
    Michael~Isard and
    Yangqing Jia and
    Rafal~Jozefowicz and
    Lukasz~Kaiser and
    Manjunath~Kudlur and
    Josh~Levenberg and
    Dandelion~Man\'{e} and
    Rajat~Monga and
    Sherry~Moore and
    Derek~Murray and
    Chris~Olah and
    Mike~Schuster and
    Jonathon~Shlens and
    Benoit~Steiner and
    Ilya~Sutskever and
    Kunal~Talwar and
    Paul~Tucker and
    Vincent~Vanhoucke and
    Vijay~Vasudevan and
    Fernanda~Vi\'{e}gas and
    Oriol~Vinyals and
    Pete~Warden and
    Martin~Wattenberg and
    Martin~Wicke and
    Yuan~Yu and
    Xiaoqiang~Zheng},
  year={2015},
}

@article{Laurent_Dinh_et_al_2017,
       author = {{Dinh}, Laurent and {Pascanu}, Razvan and {Bengio}, Samy and
         {Bengio}, Yoshua},
        title = "{Sharp Minima Can Generalize For Deep Nets}",
      journal = {arXiv e-prints},
         year = "2017",
        month = "Mar",
          eid = {arXiv:1703.04933},
archivePrefix = {arXiv},
       eprint = {1703.04933},
 primaryClass = {cs.LG},
       adsurl = {https://ui.adsabs.harvard.edu/abs/2017arXiv170304933D}
}

@article{R._Linsker1988,
 author  = {R. Linsker},
 title  = {Self-organization in a perceptual network},
 journaltitle = {Computer},
 year  = {1988},
 month  = {3},
 publisher = {Institute of Electrical and Electronics Engineers (IEEE)},
 volume  = {21},
 number  = {3},
 doi  = {10.1109/2.36},
 url  = {http://dx.doi.org/10.1109/2.36},
 pages  = {105-117},
 issn  = {0018-9162},
}

@techreport{CIFAR,
 title  = {Learning multiple layers of features from tiny images. },
 author  = {Alex Krizhevsky and
      Geoffrey Hinton},
 year  = {2009},
 institution = {Department of Computer Science, University of Toronto},
}

@article{MNIST, author	= {Y. Lecun and L. Bottou and Y. Bengio and P. Haffner}, title	= {Gradient-based learning applied to document recognition}, journaltitle	= {Proceedings of the IEEE}, year	= {1998}, publisher	= {Institute of Electrical and Electronics Engineers (IEEE)}, volume	= {86}, number	= {11}, doi	= {10.1109/5.726791}, url = {http://dx.doi.org/10.1109/5.726791}, pages	= {2278-2324}, issn	= {0018-9219}, }

@article{lecun2010mnist,
  title={MNIST handwritten digit database},
  author={LeCun, Yann and Cortes, Corinna and Burges, CJ},
  journal={ATT Labs [Online]. Available: http://yann. lecun. com/exdb/mnist},
  volume={2},
  year={2010}
}

@ARTICLE{Srivastava_dropout,
    author = {Nitish Srivastava and Geoffrey Hinton and Alex Krizhevsky and Ilya Sutskever and Ruslan Salakhutdinov and Yoshua Bengio},
    title = {Dropout: A simple way to prevent neural networks from overfitting},
    journal = {The Journal of Machine Learning Research},
    year = {},
    pages = {2014}
}

@article{L2Noise,
 author = {Bishop, Chris M.},
 title = {Training with Noise is Equivalent to Tikhonov Regularization},
 journal = {Neural Comput.},
 issue_date = {Jan. 1995},
 volume = {7},
 number = {1},
 month = jan,
 year = {1995},
 issn = {0899-7667},
 pages = {108--116},
 numpages = {9},
 url = {http://dx.doi.org/10.1162/neco.1995.7.1.108},
 doi = {10.1162/neco.1995.7.1.108},
 acmid = {211185},
 publisher = {MIT Press},
 address = {Cambridge, MA, USA},
}

@inproceedings{ExploringGeneralization,
author = {Neyshabur, Behnam and Bhojanapalli, Srinadh and McAllester, David and Srebro, Nathan},
title = {Exploring Generalization in Deep Learning},
year = {2017},
isbn = {9781510860964},
publisher = {Curran Associates Inc.},
address = {Red Hook, NY, USA},
booktitle = {Proceedings of the 31st International Conference on Neural Information Processing Systems},
pages = {5949–5958},
numpages = {10},
location = {Long Beach, California, USA},
series = {NIPS’17}
}

@ARTICLE{2017arXiv171005468K,
       author = {{Kawaguchi}, Kenji and {Pack Kaelbling}, Leslie and {Bengio}, Yoshua},
        title = "{Generalization in Deep Learning}",
      journal = {arXiv e-prints},
         year = "2017",
        month = "Oct",
          eid = {arXiv:1710.05468},
        pages = {arXiv:1710.05468},
archivePrefix = {arXiv},
       eprint = {1710.05468},
 primaryClass = {stat.ML},
       adsurl = {https://ui.adsabs.harvard.edu/abs/2017arXiv171005468K}
}

@ARTICLE{adam,
       author = {{Kingma}, Diederik P. and {Ba}, Jimmy},
        title = "{Adam: A Method for Stochastic Optimization}",
      journal = {arXiv e-prints},
         year = "2014",
        month = "Dec",
          eid = {arXiv:1412.6980},
        pages = {arXiv:1412.6980},
archivePrefix = {arXiv},
       eprint = {1412.6980},
 primaryClass = {cs.LG},
       adsurl = {https://ui.adsabs.harvard.edu/abs/2014arXiv1412.6980K}
}

@article{deepobs,
  author    = {Frank Schneider and
               Lukas Balles and
               Philipp Hennig},
  title     = {DeepOBS: {A} Deep Learning Optimizer Benchmark Suite},
  journal   = {CoRR},
  volume    = {abs/1903.05499},
  year      = {2019},
  url       = {http://arxiv.org/abs/1903.05499},
  eprinttype = {arXiv},
  eprint    = {1903.05499},
  bibsource = {dblp computer science bibliography, https://dblp.org}
}

@InProceedings{crowdedvalley,
  title = 	 {Descending through a Crowded Valley - Benchmarking Deep Learning Optimizers},
  author =       {Schmidt, Robin M and Schneider, Frank and Hennig, Philipp},
  booktitle = 	 {Proceedings of the 38th International Conference on Machine Learning},
  pages = 	 {9367--9376},
  year = 	 {2021},
  editor = 	 {Meila, Marina and Zhang, Tong},
  volume = 	 {139},
  series = 	 {Proceedings of Machine Learning Research},
  month = 	 {18--24 Jul},
  publisher =    {PMLR},
  url = 	 {http://proceedings.mlr.press/v139/schmidt21a.html}
}

@article{EntropySGD,
  author    = {Pratik Chaudhari and
               Anna Choromanska and
               Stefano Soatto and
               Yann LeCun and
               Carlo Baldassi and
               Christian Borgs and
               Jennifer T. Chayes and
               Levent Sagun and
               Riccardo Zecchina},
  title     = {Entropy-SGD: Biasing Gradient Descent Into Wide Valleys},
  journal   = {CoRR},
  volume    = {abs/1611.01838},
  year      = {2016},
  url       = {http://arxiv.org/abs/1611.01838},
  eprinttype = {arXiv},
  eprint    = {1611.01838},
  bibsource = {dblp computer science bibliography, https://dblp.org}
}

@article{goh2017why,
  author = {Goh, Gabriel},
  title = {Why Momentum Really Works},
  journal = {Distill},
  year = {2017},
  url = {http://distill.pub/2017/momentum},
  doi = {10.23915/distill.00006}
}

@article{JMLR:v17:15-084,
  author  = {Weijie Su and Stephen Boyd and Emmanuel J. Cand{{\`e}}s},
  title   = {A Differential Equation for Modeling Nesterov's Accelerated Gradient Method: Theory and Insights},
  journal = {Journal of Machine Learning Research},
  year    = {2016},
  volume  = {17},
  number  = {153},
  pages   = {1--43},
  url     = {http://jmlr.org/papers/v17/15-084.html}
}

@article{smith_cyclic,
  author    = {Leslie N. Smith},
  title     = {Cyclical Learning Rates for Training Neural Networks},
  journal   = {CoRR},
  volume    = {abs/1506.01186},
  year      = {2015},
  url       = {http://arxiv.org/abs/1506.01186},
  eprinttype = {arXiv},
  eprint    = {1506.01186},
  bibsource = {dblp computer science bibliography, https://dblp.org}
}

@article{superconvergence,
  author    = {Leslie N. Smith and
               Nicholay Topin},
  title     = {Super-Convergence: Very Fast Training of Residual Networks Using Large
               Learning Rates},
  journal   = {CoRR},
  volume    = {abs/1708.07120},
  year      = {2017},
  url       = {http://arxiv.org/abs/1708.07120},
  eprinttype = {arXiv},
  eprint    = {1708.07120},
  bibsource = {dblp computer science bibliography, https://dblp.org}
}

@article{matchedfilter_perspective,
  author    = {Ljubisa Stankovic and
               Danilo P. Mandic},
  title     = {Convolutional Neural Networks Demystified: {A} Matched Filtering Perspective
               Based Tutorial},
  journal   = {CoRR},
  volume    = {abs/2108.11663},
  year      = {2021},
  url       = {https://arxiv.org/abs/2108.11663},
  eprinttype = {arXiv},
  eprint    = {2108.11663},
  bibsource = {dblp computer science bibliography, https://dblp.org}
}

@book{mlrefined, place={Cambridge}, title={Machine Learning Refined: Foundations, Algorithms, and Applications}, DOI={10.1017/CBO9781316402276}, publisher={Cambridge University Press}, author={Watt, Jeremy and Borhani, Reza and Katsaggelos, Aggelos K.}, year={2016}}

@article{pohlen2018observe,
  title={Observe and look further: Achieving consistent performance on atari},
  author={Pohlen, Tobias and Piot, Bilal and Hester, Todd and Azar, Mohammad Gheshlaghi and Horgan, Dan and Budden, David and Barth-Maron, Gabriel and van Hasselt, Hado and Quan, John and Ve{\v{c}}er{\'\i}k, Mel and others},
  journal={arXiv preprint arXiv:1805.11593},
  year={2018}
}

@article{muzero,
  author    = {Julian Schrittwieser and
               Ioannis Antonoglou and
               Thomas Hubert and
               Karen Simonyan and
               Laurent Sifre and
               Simon Schmitt and
               Arthur Guez and
               Edward Lockhart and
               Demis Hassabis and
               Thore Graepel and
               Timothy P. Lillicrap and
               David Silver},
  title     = {Mastering Atari, Go, Chess and Shogi by Planning with a Learned Model},
  journal   = {CoRR},
  volume    = {abs/1911.08265},
  year      = {2019},
  url       = {http://arxiv.org/abs/1911.08265},
  eprinttype = {arXiv},
  eprint    = {1911.08265},
  bibsource = {dblp computer science bibliography, https://dblp.org}
}

@Article{Robbins1951,
  author    = {Robbins, Herbert and Monro, Sutton},
  journal   = {The Annals of Mathematical Statistics},
  title     = {{A Stochastic Approximation Method}},
  year      = {1951},
  number    = {3},
  oppages     = {400--407},
  volume    = {22},
  optdoi    = {10.1214/aoms/1177729586},
  optissn   = {0003-4851},
  optmonth  = {sep},
  publisher = {Institute of Mathematical Statistics},
}

@Article{Polyak1964,
  author   = {Polyak, B. T.},
  journal  = {USSR Computational Mathematics and Mathematical Physics},
  title    = {{Some methods of speeding up the convergence of iteration methods}},
  year     = {1964},
  number   = {5},
  oppages    = {1--17},
  volume   = {4},
  optdoi   = {10.1016/0041-5553(64)90137-5},
  optissn  = {00415553},
}

@Article{Nesterov1983,
  author  = {Nesterov, Yurii},
  journal = {Soviet Mathematics Doklady},
  title   = {{A method for solving the convex programming problem with convergence rate $O(1/k^2)$}},
  year    = {1983},
  volume  = {27},
  oppages = {372--376},
}

@inproceedings{Kingma2015,
  author    = {Diederik P. Kingma and
               Jimmy Ba},
  title     = {{Adam: A Method for Stochastic Optimization}},
  booktitle = {3rd International Conference on Learning Representations, {ICLR}},
  year      = {2015}
}

@misc{Tieleman2012,
author = {Tieleman, Tijmen and Hinton, Geoffrey},
publisher = {COURSERA: Neural Networks for Machine Learning},
title = {{Lecture 6.5---RMSProp: Divide the gradient by a running average of its recent magnitude}},
year = {2012}
}

@Article{Zhang2019a,
  author        = {Zhang, Michael R. and Lucas, James and Hinton, Geoffrey and Ba, Jimmy},
  journal       = {Advances in Neural Information Processing Systems 32, NeurIPS},
  title         = {{Lookahead Optimizer: k steps forward, 1 step back}},
  year          = {2019},
  archiveprefix = {arXiv},
  arxivid       = {1907.08610},
  eprint        = {1907.08610},
  optmonth      = {jul},
  opturl        = {http://arxiv.org/abs/1907.08610},
}

@inproceedings{Dozat2016IncorporatingNM,
  title={{Incorporating Nesterov Momentum into Adam}},
  author={Timothy Dozat},
  booktitle = {4th International Conference on Learning Representations, {ICLR}},
  year={2016}
}

@InProceedings{Reddi2018,
  author        = {Reddi, Sashank J. and Kale, Satyen and Kumar, Sanjiv},
  booktitle     = {6th International Conference on Learning Representations, ICLR},
  title         = {{On the Convergence of Adam and Beyond}},
  year          = {2018},
  oppages         = {1--23},
  archiveprefix = {arXiv},
  arxivid       = {1904.09237},
  eprint        = {1904.09237},
  opturl        = {http://arxiv.org/abs/1904.09237},
}

@InProceedings{Luo2019,
  author        = {Luo, Liangchen and Xiong, Yuanhao and Liu, Yan and Sun, Xu},
  booktitle     = {7th International Conference on Learning Representations, ICLR},
  title         = {{Adaptive Gradient Methods with Dynamic Bound of Learning Rate}},
  year          = {2019},
  archiveprefix = {arXiv},
  arxivid       = {1902.09843},
  eprint        = {1902.09843},
  optmonth      = {feb},
  opturl        = {http://arxiv.org/abs/1902.09843},
}

@article{Ruder16,
  author    = {Sebastian Ruder},
  title     = {An overview of gradient descent optimization algorithms},
  journal   = {CoRR},
  volume    = {abs/1609.04747},
  year      = {2016},
  url       = {http://arxiv.org/abs/1609.04747},
  eprinttype = {arXiv},
  eprint    = {1609.04747},
  bibsource = {dblp computer science bibliography, https://dblp.org}
}

@book{DLrefined,  place={Cambridge}, edition={2}, title={Machine Learning Refined: Foundations, Algorithms, and Applications}, DOI={10.1017/9781108690935}, publisher={Cambridge University Press}, author={Watt, Jeremy and Borhani, Reza and Katsaggelos, Aggelos K.}, year={2020}}

@article{fmnist,
  author    = {Han Xiao and
               Kashif Rasul and
               Roland Vollgraf},
  title     = {Fashion-MNIST: a Novel Image Dataset for Benchmarking Machine Learning
               Algorithms},
  journal   = {CoRR},
  volume    = {abs/1708.07747},
  year      = {2017},
  url       = {http://arxiv.org/abs/1708.07747},
  eprinttype = {arXiv},
  eprint    = {1708.07747},
  bibsource = {dblp computer science bibliography, https://dblp.org}
}

@inproceedings{svhn,
title	= {Reading Digits in Natural Images with Unsupervised Feature Learning},
author	= {Yuval Netzer and Tao Wang and Adam Coates and Alessandro Bissacco and Bo Wu and Andrew Y. Ng},
year	= {2011},
URL	= {http://ufldl.stanford.edu/housenumbers/nips2011_housenumbers.pdf},
booktitle	= {NIPS Workshop on Deep Learning and Unsupervised Feature Learning 2011}
}

@article{wrn,
  author    = {Sergey Zagoruyko and
               Nikos Komodakis},
  title     = {Wide Residual Networks},
  journal   = {CoRR},
  volume    = {abs/1605.07146},
  year      = {2016},
  url       = {http://arxiv.org/abs/1605.07146},
  eprinttype = {arXiv},
  eprint    = {1605.07146},
  bibsource = {dblp computer science bibliography, https://dblp.org}
}

@ARTICLE{allcnn,
       author = {{Springenberg}, Jost Tobias and {Dosovitskiy}, Alexey and {Brox}, Thomas and {Riedmiller}, Martin},
        title = "{Striving for Simplicity: The All Convolutional Net}",
      journal = {arXiv e-prints},
         year = 2014,
        month = dec,
          eid = {arXiv:1412.6806},
        pages = {arXiv:1412.6806},
archivePrefix = {arXiv},
       eprint = {1412.6806},
 primaryClass = {cs.LG},
       adsurl = {https://ui.adsabs.harvard.edu/abs/2014arXiv1412.6806S}
}

@article{biaswidevalleys,
  author    = {Pratik Chaudhari and
               Anna Choromanska and
               Stefano Soatto and
               Yann LeCun and
               Carlo Baldassi and
               Christian Borgs and
               Jennifer T. Chayes and
               Levent Sagun and
               Riccardo Zecchina},
  title     = {Entropy-SGD: Biasing Gradient Descent Into Wide Valleys},
  journal   = {CoRR},
  volume    = {abs/1611.01838},
  year      = {2016},
  url       = {http://arxiv.org/abs/1611.01838},
  eprinttype = {arXiv},
  eprint    = {1611.01838},
  bibsource = {dblp computer science bibliography, https://dblp.org}
}

@misc{flatminima_comparison,
      title={A Fair Comparison of Two Popular Flat Minima Optimizers: Stochastic Weight Averaging vs. Sharpness-Aware Minimization}, 
      author={Jean Kaddour and Linqing Liu and Ricardo Silva and Matt J. Kusner},
      year={2022},
      eprint={2202.00661},
      archivePrefix={arXiv},
      primaryClass={cs.LG}
}

@misc{
flatminima_largemargins,
title={On Flat Minima, Large Margins and Generalizability},
author={Daniel Lengyel and Nicholas Jennings and Panos Parpas and Nicholas Kantas},
year={2021},
url={https://openreview.net/forum?id=Ki5Mv0iY8C}
}

@article{sharpness_aware,
  author    = {Pierre Foret and
               Ariel Kleiner and
               Hossein Mobahi and
               Behnam Neyshabur},
  title     = {Sharpness-Aware Minimization for Efficiently Improving Generalization},
  journal   = {CoRR},
  volume    = {abs/2010.01412},
  year      = {2020},
  url       = {https://arxiv.org/abs/2010.01412},
  eprinttype = {arXiv},
  eprint    = {2010.01412},
  bibsource = {dblp computer science bibliography, https://dblp.org}
}

@article{asymmetric_flatminima,
  author    = {Haowei He and
               Gao Huang and
               Yang Yuan},
  title     = {Asymmetric Valleys: Beyond Sharp and Flat Local Minima},
  journal   = {CoRR},
  volume    = {abs/1902.00744},
  year      = {2019},
  url       = {http://arxiv.org/abs/1902.00744},
  eprinttype = {arXiv},
  eprint    = {1902.00744},
  bibsource = {dblp computer science bibliography, https://dblp.org}
}

@article{fantastic_measures,
  author    = {Yiding Jiang and
               Behnam Neyshabur and
               Hossein Mobahi and
               Dilip Krishnan and
               Samy Bengio},
  title     = {Fantastic Generalization Measures and Where to Find Them},
  journal   = {CoRR},
  volume    = {abs/1912.02178},
  year      = {2019},
  url       = {http://arxiv.org/abs/1912.02178},
  eprinttype = {arXiv},
  eprint    = {1912.02178},
  bibsource = {dblp computer science bibliography, https://dblp.org}
}

@misc{natekar2020representation,
      title={Representation Based Complexity Measures for Predicting Generalization in Deep Learning}, 
      author={Parth Natekar and Manik Sharma},
      year={2020},
      eprint={2012.02775},
      archivePrefix={arXiv},
      primaryClass={cs.LG}
}

@article{predicting_generalisation_via_noise,
  author    = {Depen Morwani and
               Rahul Vashisht and
               Harish G. Ramaswamy},
  title     = {Using noise resilience for ranking generalization of deep neural networks},
  journal   = {CoRR},
  volume    = {abs/2012.08854},
  year      = {2020},
  url       = {https://arxiv.org/abs/2012.08854},
  eprinttype = {arXiv},
  eprint    = {2012.08854},
  bibsource = {dblp computer science bibliography, https://dblp.org}
}

@article{backprop,
  author = {Rumelhart, David E. and Hinton, Geoffrey E. and Williams, Ronald J.},
  doi = {10.1038/323533a0},
  journal = {Nature},
  number = 6088,
  pages = {533--536},
  title = {{Learning Representations by Back-propagating Errors}},
  url = {http://www.nature.com/articles/323533a0},
  volume = 323,
  year = 1986
}

@InProceedings{uniqueproperties_flatminima,
  title = 	 {Unique Properties of Flat Minima in Deep Networks},
  author =       {Mulayoff, Rotem and Michaeli, Tomer},
  booktitle = 	 {Proceedings of the 37th International Conference on Machine Learning},
  pages = 	 {7108--7118},
  year = 	 {2020},
  editor = 	 {III, Hal Daumé and Singh, Aarti},
  volume = 	 {119},
  series = 	 {Proceedings of Machine Learning Research},
  month = 	 {13--18 Jul},
  publisher =    {PMLR},
  url = 	 {https://proceedings.mlr.press/v119/mulayoff20a.html}
}

@article{powersign,
  author    = {Irwan Bello and
               Barret Zoph and
               Vijay Vasudevan and
               Quoc V. Le},
  title     = {Neural Optimizer Search with Reinforcement Learning},
  journal   = {CoRR},
  volume    = {abs/1709.07417},
  year      = {2017},
  url       = {http://arxiv.org/abs/1709.07417},
  eprinttype = {arXiv},
  eprint    = {1709.07417},
  bibsource = {dblp computer science bibliography, https://dblp.org}
}

@article{difficultytraininglanguage,
  author    = {Razvan Pascanu and
               Tom{\'{a}}s Mikolov and
               Yoshua Bengio},
  title     = {Understanding the exploding gradient problem},
  journal   = {CoRR},
  volume    = {abs/1211.5063},
  year      = {2012},
  url       = {http://arxiv.org/abs/1211.5063},
  eprinttype = {arXiv},
  eprint    = {1211.5063},
  bibsource = {dblp computer science bibliography, https://dblp.org}
}

@book{Mikolov,
author = {T. Mikolov},
title = {Statistical Language Models based on Neural Networks},
year = {2012},
publisher = {Ph.D. thesis, Brno University of Technology}
}

@misc{chollet2015keras,
  title={Keras},
  author={Chollet, Fran\c{c}ois and others},
  year={2015},
  howpublished={\url{https://keras.io}},
}

@article{chiyuan_still,
author = {Zhang, Chiyuan and Bengio, Samy and Hardt, Moritz and Recht, Benjamin and Vinyals, Oriol},
title = {Understanding Deep Learning (Still) Requires Rethinking Generalization},
year = {2021},
issue_date = {March 2021},
publisher = {Association for Computing Machinery},
address = {New York, NY, USA},
volume = {64},
number = {3},
issn = {0001-0782},
url = {https://doi.org/10.1145/3446776},
doi = {10.1145/3446776},
journal = {Commun. ACM},
month = {feb},
pages = {107–115},
numpages = {9}
}

@article{increase_batch_size,
  author    = {Samuel L. Smith and
               Pieter{-}Jan Kindermans and
               Quoc V. Le},
  title     = {Don't Decay the Learning Rate, Increase the Batch Size},
  journal   = {CoRR},
  volume    = {abs/1711.00489},
  year      = {2017},
  url       = {http://arxiv.org/abs/1711.00489},
  eprinttype = {arXiv},
  eprint    = {1711.00489},
  bibsource = {dblp computer science bibliography, https://dblp.org}
}
\newpage

\appendix

\section{Toy Model}\label{sec.toy}

In this brief appendix, we consider a toy models that give some intuition related to points discussed in the main text. \par

\subsection{Single node}

\begin{figure}[!htbp]
 
    \centering
    \includegraphics[width=0.65\linewidth]{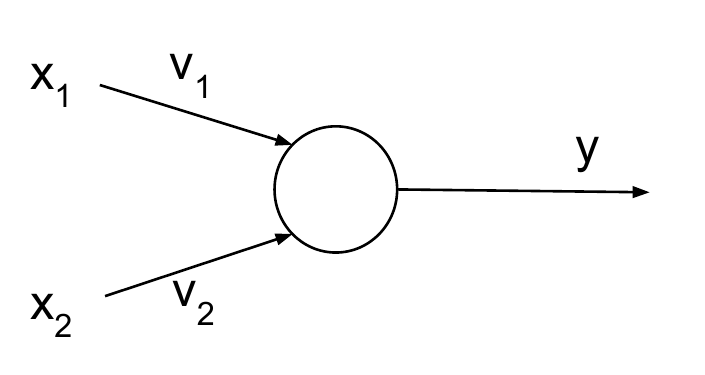}
  
    \caption{Toy model: single node }
    \label{fig:toy1}
\end{figure}

First, consider a single node model (figure \ref{fig:toy1}), with two inputs that have normally distribution samples $N(a,\sigma^2)$ and a constant target, $a$. The output will have the distribution $N\left((v_1 + v_2)a,\sigma^2(v_1^2 + v_2^2)\right)$. Clearly, the solution is any point on the line $v_1 + v_2 = 1$. The standard deviation of the error is minimised when $v_1 = v_2 = a/2$, hence this is the weight combination that is `SNR optimal'. (Of course, in this example, SNR-optimal is simply the MSE solution.)\par 

\begin{figure}[!htbp]
 
    \centering
    \includegraphics[width=0.9\linewidth]{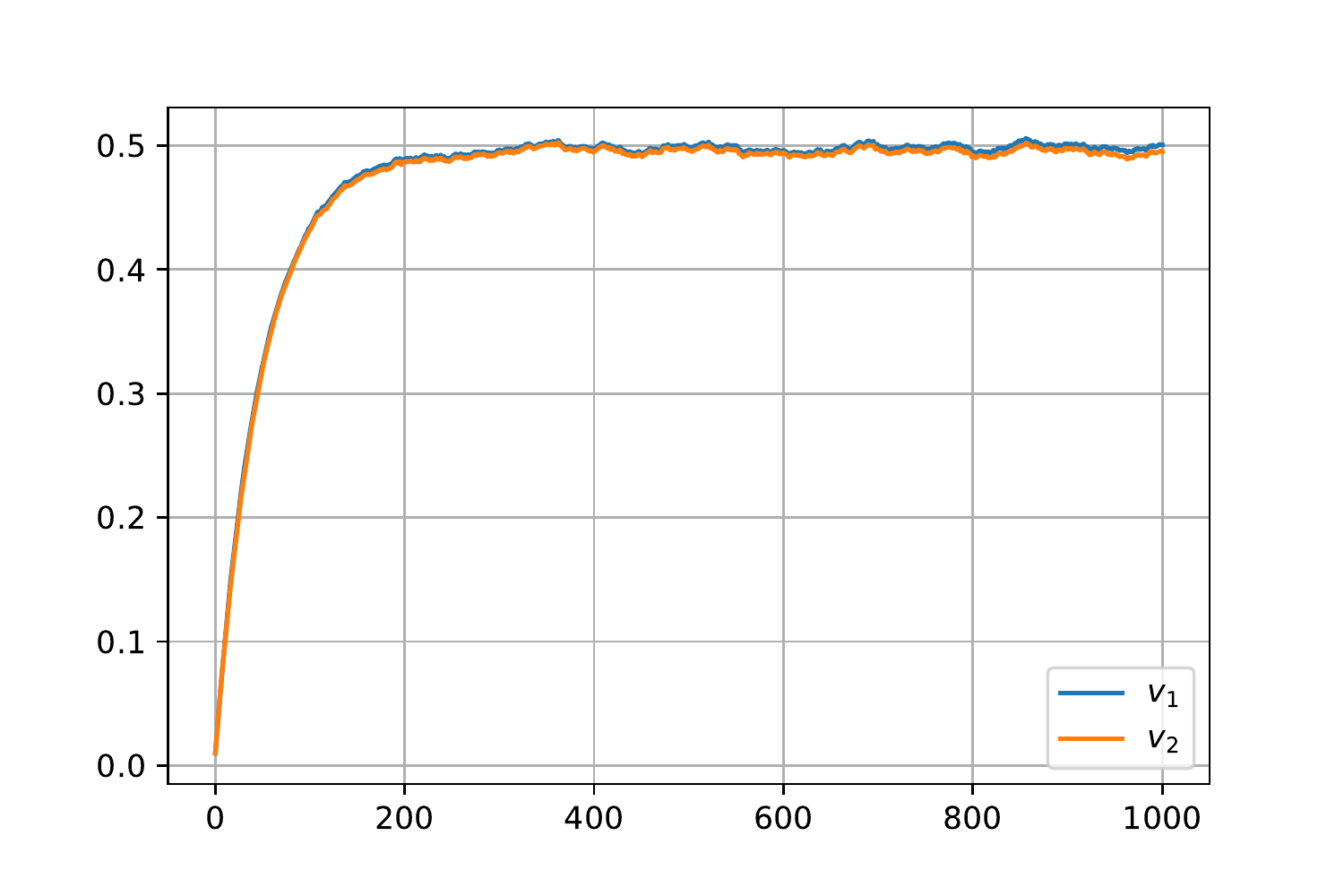}
    \caption{Results for $v_{1} = 0.01, v_{2} = 0.01$ }
    \label{fig:toyresults_v1}
\end{figure}

\begin{figure}[!htbp]
    \centering
    \begin{subfigure}[b]{0.90\linewidth}
       \includegraphics[width=\linewidth]{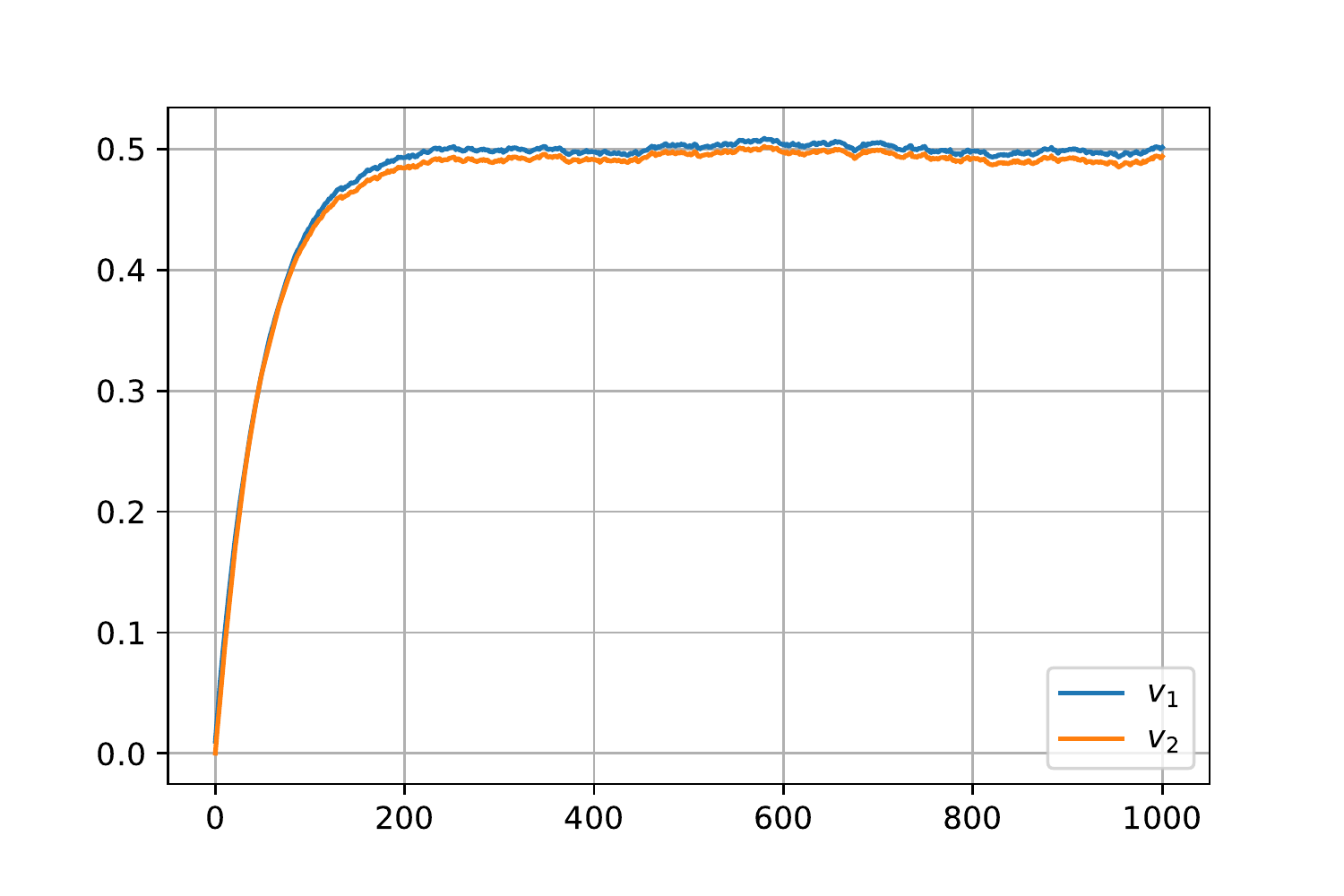}
        \caption{initial convergence}
    \end{subfigure}

    \par
    \medskip
     
    \begin{subfigure}[b]{0.90\linewidth}
        \includegraphics[width=\linewidth]{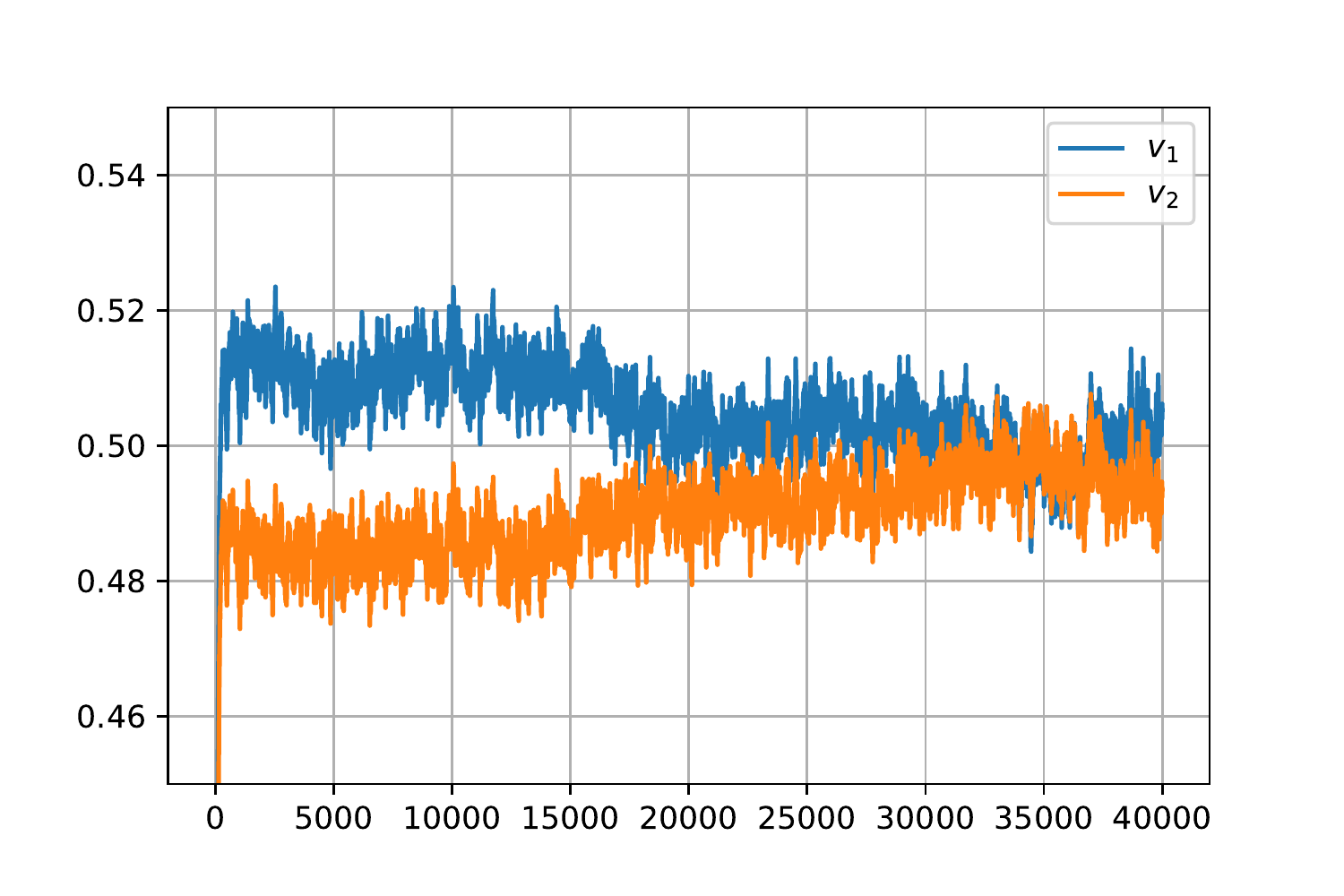}
        \caption{training, zoomed}
    \end{subfigure}

    \caption{Results for $v_{1} = 0.01, v_{2} = 0.0001$ }
    \label{fig:toyresults_v2}
\end{figure}

 Figures \ref{fig:toyresults_v1} and \ref{fig:toyresults_v2} show typical examples of training this model with a loss $L = \left(v_1 x_1 + v_2 x_2 - a\right)^2$. In the initial stages, the model converge to a point on $v_1 + v_2 = 1$, but one that maintains where $v_1 - v_2$ is close to its initialised value. Eventually, the regularisation effect of the inherent noise (see \cite{L2Noise}) pushes the network to an SNR-optimal solution. These are, of course, simple examples, but act as reference points for the following discussion.\par

\newpage

\subsection{Three node model}
     
\begin{figure}[!htbp]
 
    \centering
    \includegraphics[width=0.9\linewidth]{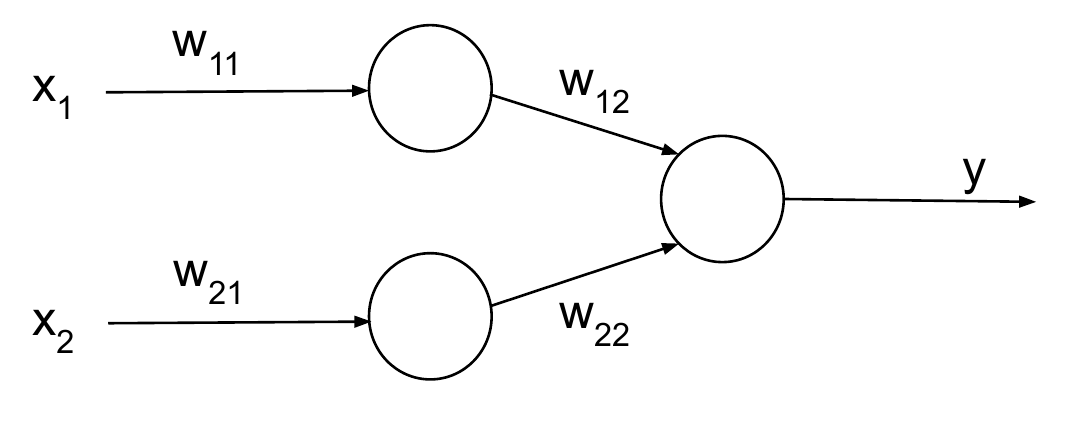}

    \caption{Toy model: three nodes }
    \label{fig:toy2}
\end{figure}

We now consider a three-node model shown in  figure \ref{fig:toy2}. It is easy to see that as feed-forward functions there is no real difference between (a) and (b). In fact, if $v_i$=$w_{i1}w_{i2}$ then they are essentially identical. \par

However, the behaviour under SGD/backprop is notably different. To see this, let’s consider the total change in the paths from each input after a single iteration. For the two-layer function, each path is characterised by the product of the two weights. (As already noted this product has an identical role to the single weights in the 1-layer model.) After the first back-propagation iteration the products will be:
\begin{align*}
    w_{11}w_{12} &\leftarrow w_{11}w_{12}\left(1+2S_1\right)^2-S_1\left(w_{11}^2+w_{12}^2\right) \\
        w_{21}w_{22} &\leftarrow w_{21}w_{22}\left(1+2S_2\right)^2-S_2\left(w_{21}^2+w_{22}^2\right) 
\end{align*}
where $S_j =\sum_n{x_j^{(n)}\left(y^{(n)}-a\right)}$.

We see that even if the products $w_{i1}w_{i2}$ are identical at the start and $S_1\sim{}S_2$, any difference in the way the weights are distributed can cause the two products to diverge and hence the \snr{} to degrade. \par

Let’s consider a case where the SNR is good at the start of the training process, with $w_{11}w_{12} = w_{21}w_{22}$, but with 
\begin{align*}
w_{12} &= w_{11}\\
w_{21} &= w_{11} \kappa \\
w_{22} &= w_{11}/\kappa    
\end{align*}
for some $\kappa$.\par

When we converge, $w_{11}w_{12}$ will no longer be close to $w_{21}w_{22}$. Hence, the SNR of the 2-layer model will have degraded and we will see greater error in individual results (even if the mean error remains zero).\par

  \begin{figure}[!htbp]
 
   \centering
  
    \begin{subfigure}[b]{0.90\linewidth}
       \includegraphics[width=\linewidth]{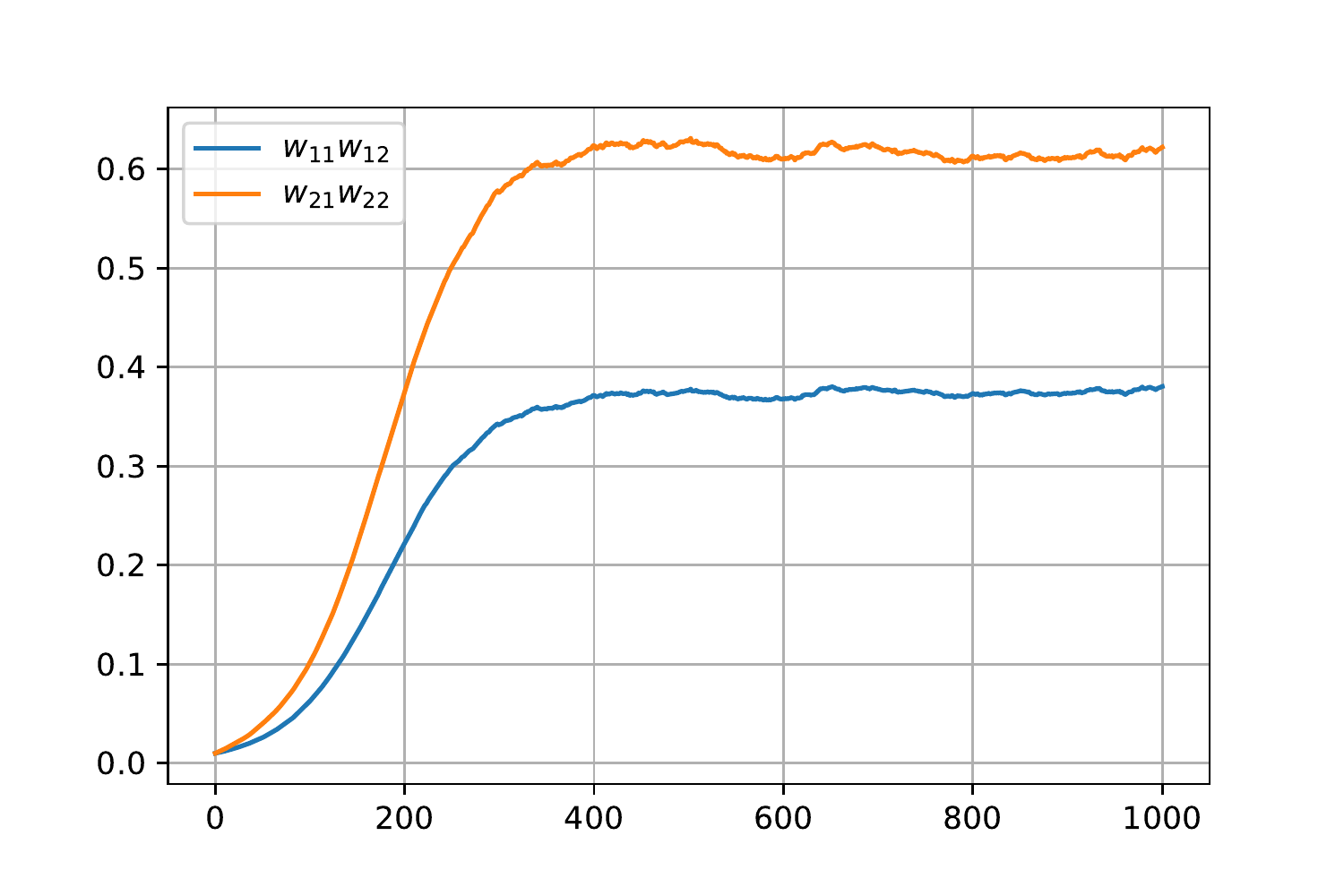}
       \caption{Gradient Descent}
     \end{subfigure}
     
        \par
        \medskip
     \begin{subfigure}[b]{0.90\linewidth}
        \includegraphics[width=\linewidth]{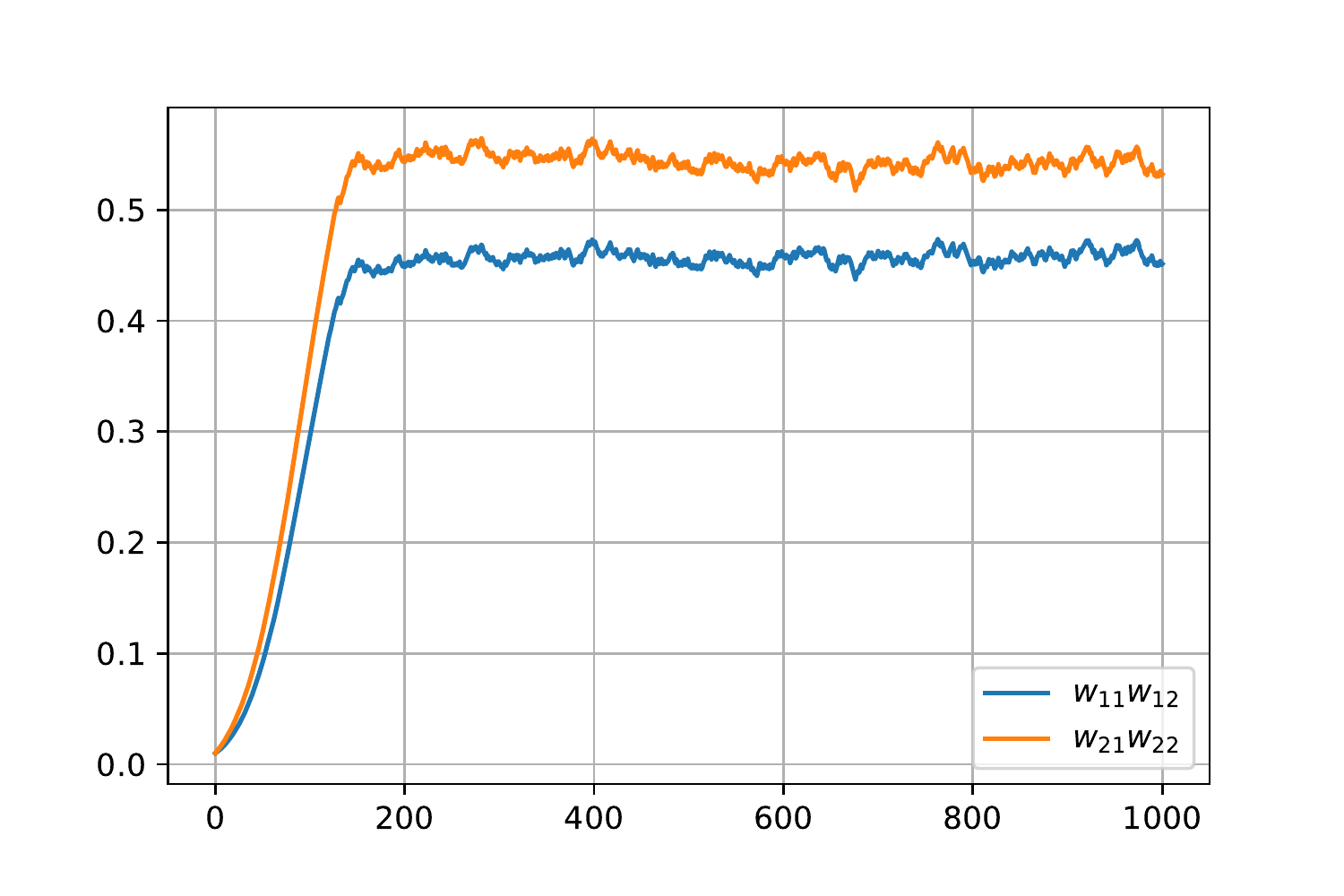}
       \caption{\nl{}}

     \end{subfigure}
  
\caption{Results for $w_{11} = w_{12} = 0.1$, $w_{12} = 0.05, w_{22} = 0.2$  }
     \label{fig:toyresults_w1}
   \end{figure}

Figure \ref{fig:toyresults_w1} (a) gives an example of the behaviour in such a case. The figure shows the shows the evolution of $w_{11}w_{12}$ and $w_{21}w_{22}$, which is the equivalent to $v_1$ and $v_2$ above. The weights are initialised with $\kappa = 0.5$; that is $w_{11} = w_{12} = 2$, $w_{21} = 1.6$, $w_{12} = 2.5$. This, then is identical to the initial point for \ref{fig:toyresults_v1}. \par

We see that when the model converges the SNR is degraded; the weighting on the two input paths no longer maintains the initial distance but diverges. For this simple case, the noise regularisation will eventually push the model to an SNR-optimal solution, \par

Figure \ref{fig:toyresults_w1} (b) shows the training path when we apply a signed square root to the updates, that is $\Delta w_{ij}~=~\alpha\ \sign{\left(\frac{\partial f}{\partial w_{ij}}\right)\sqrt{\frac{\partial f}{\partial w_{ij}}}}$. As argued in the main text, this NL approach has the goal that the initial convergence reaches to a point that is better from an \snr{} perspective.\par

\begin{figure}[!htbp]
 
   \centering
  
        \begin{subfigure}[b]{0.90\linewidth}
       \includegraphics[width=\linewidth]{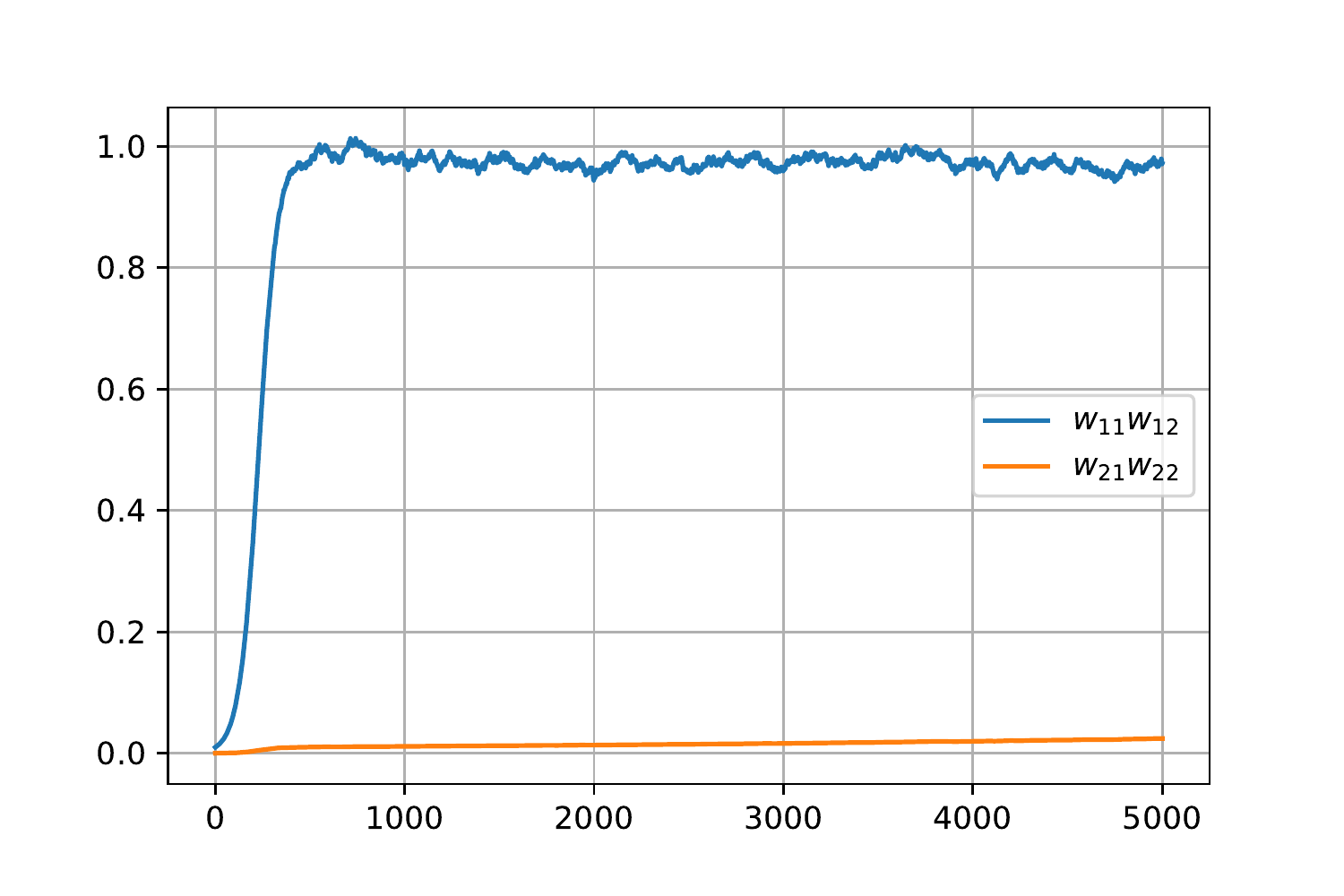}
       \caption{Gradient Descent}
     \end{subfigure}
     
        \par
        \medskip
     \begin{subfigure}[b]{0.90\linewidth}
        \includegraphics[width=\linewidth]{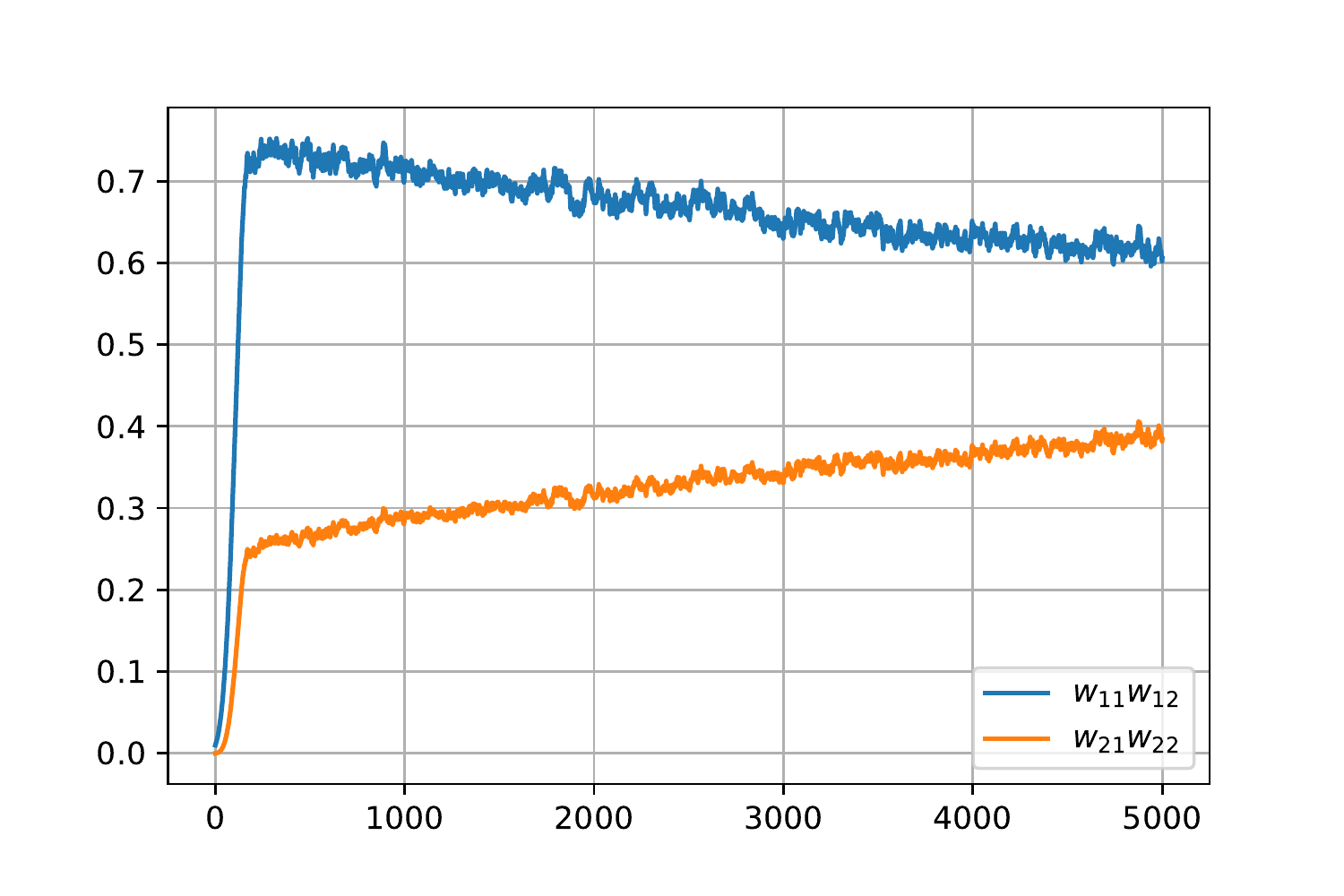}
       \caption{\nl{}}

     \end{subfigure}
  
\caption{Results for $w_{11} = w_{12} = 0.1$, $w_{12} = w_{22} = 0.01$  }
     \label{fig:toyresults_w2}
   \end{figure}

Figure \ref{fig:toyresults_w2} shows the evolution of $w_{11}w_{12}$ and $w_{21}w_{22}$, when starting with weights $w_{11} = w_{12} = 0.1$, $w_{21} = w_{12} = 0.01$. This is the equivalent starting point to in figure \ref{fig:toyresults_v2}. Again, the initial imbalance leads to suboptimal SNR as the model trains; in this case, with one input dominating. Again, when we apply a signed square root, the SNR at convergence is improved.\par

As noted, in all these cases, the regularisation effect of the inherent noise (see \cite{L2Noise}) eventually push to an SNR-optimal solution. However, we expect that, in more complex cases, the initial convergence will become more critical to the overall performance, as demonstrated in the results above. \par

From both these examples, we see that SGD applied to a DNN can frequently lead to training trajectories that are not optimal from an SNR perspective. Frequently there is a `success-breeds-success' effect, where strongly weighted paths through the network are emphasised even when more balance between paths would give better performance. This raises the possibility that, in more complex networks, the training never reaches an SNR-optimal solution -- or does so only after a long training time. \par

\clearpage
\onecolumn
\section{Training plots}\label{sec:timeplots}
In this section we give some plots of the training over time to demonstrate the dynamics of NL algorithms compared to other options. Each plot shows one NL algorithm against SGD, Momentum and Adam. The results are over the 10 repeated training runs using the parameter set selected in the random search; the plots show the mean and standard deviation of the set. \par

\subsection{FMNIST 2C2D}
\begin{figure*}[!ht]

\centering
  \includegraphics[width=.7\textwidth]{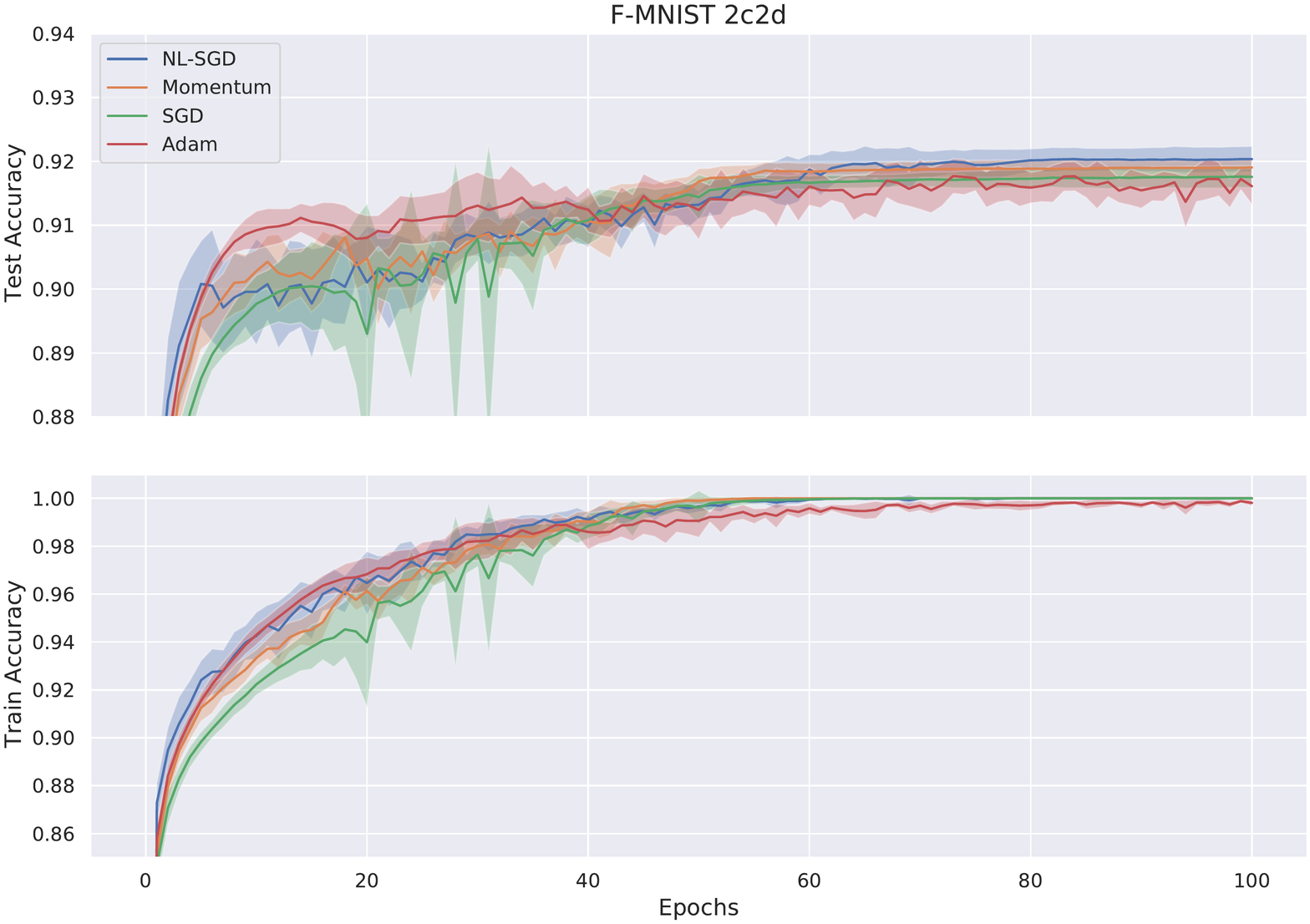}
  \caption{NL-SGD, FMNIST 2C2D (accuracy)}
  
\end{figure*}

\begin{figure*}[!ht]
\centering
  \includegraphics[width=.7\textwidth]{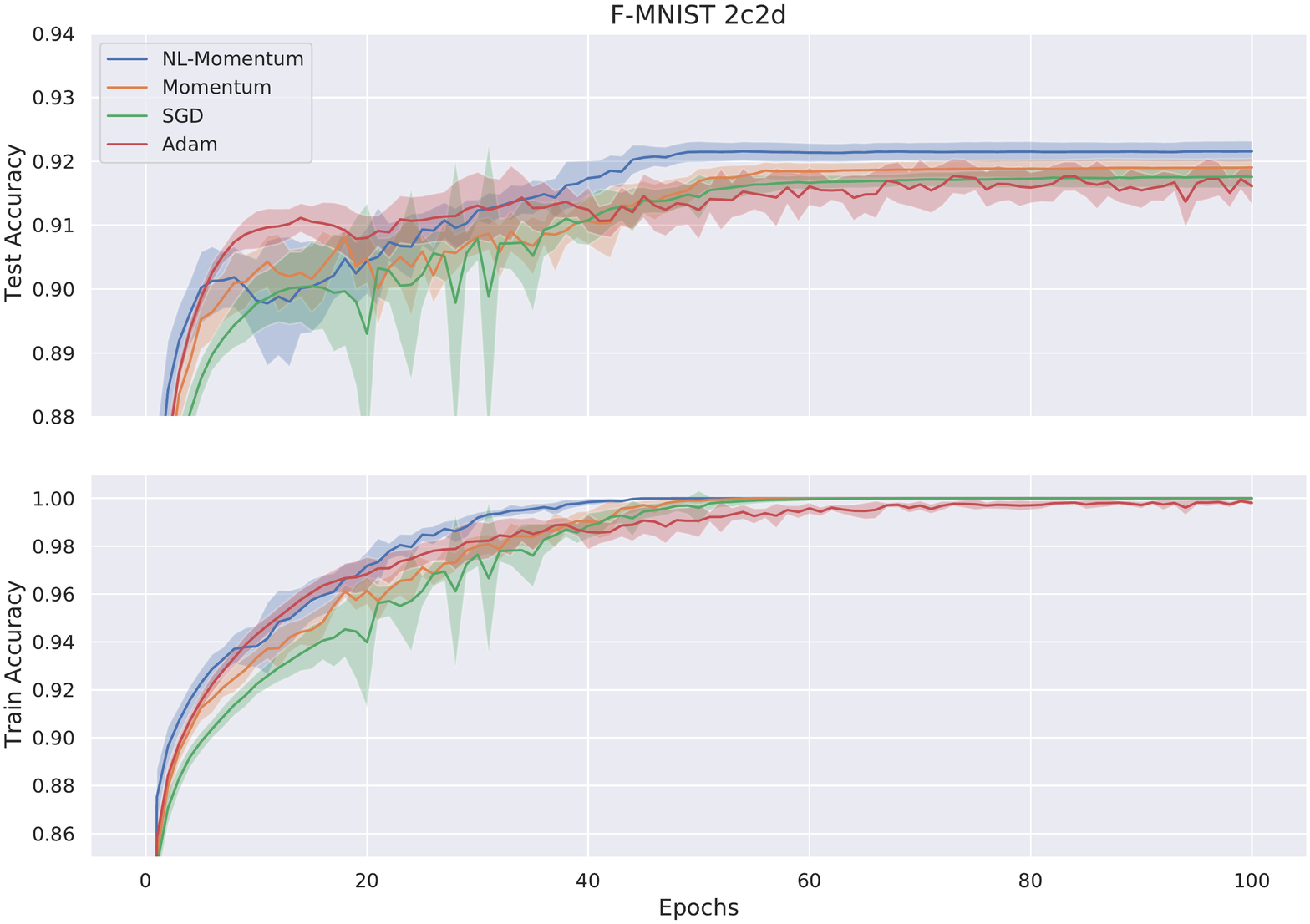}
  \caption{NL-Momentum, FMNIST 2C2D (accuracy)}
\end{figure*}

\pagebreak
\subsection{CIFAR-10 3C3D}
\begin{figure*}[!ht]

\centering
  \includegraphics[width=.7\textwidth]{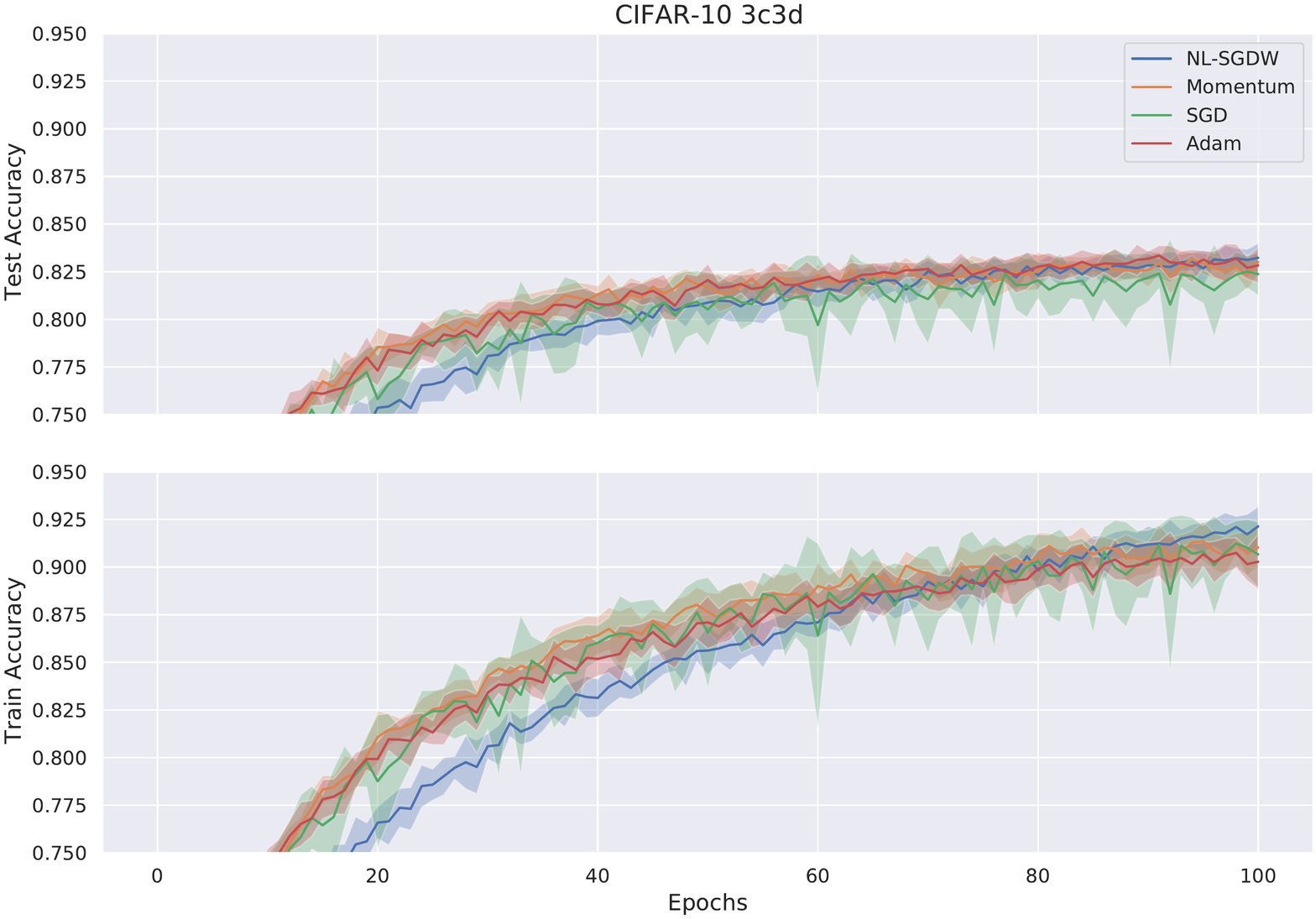}
  \caption{NL-SGD, CIFAR-10 3C3D (accuracy)}
\end{figure*}

\begin{figure*}[!ht]
\centering
  \includegraphics[width=.7\textwidth]{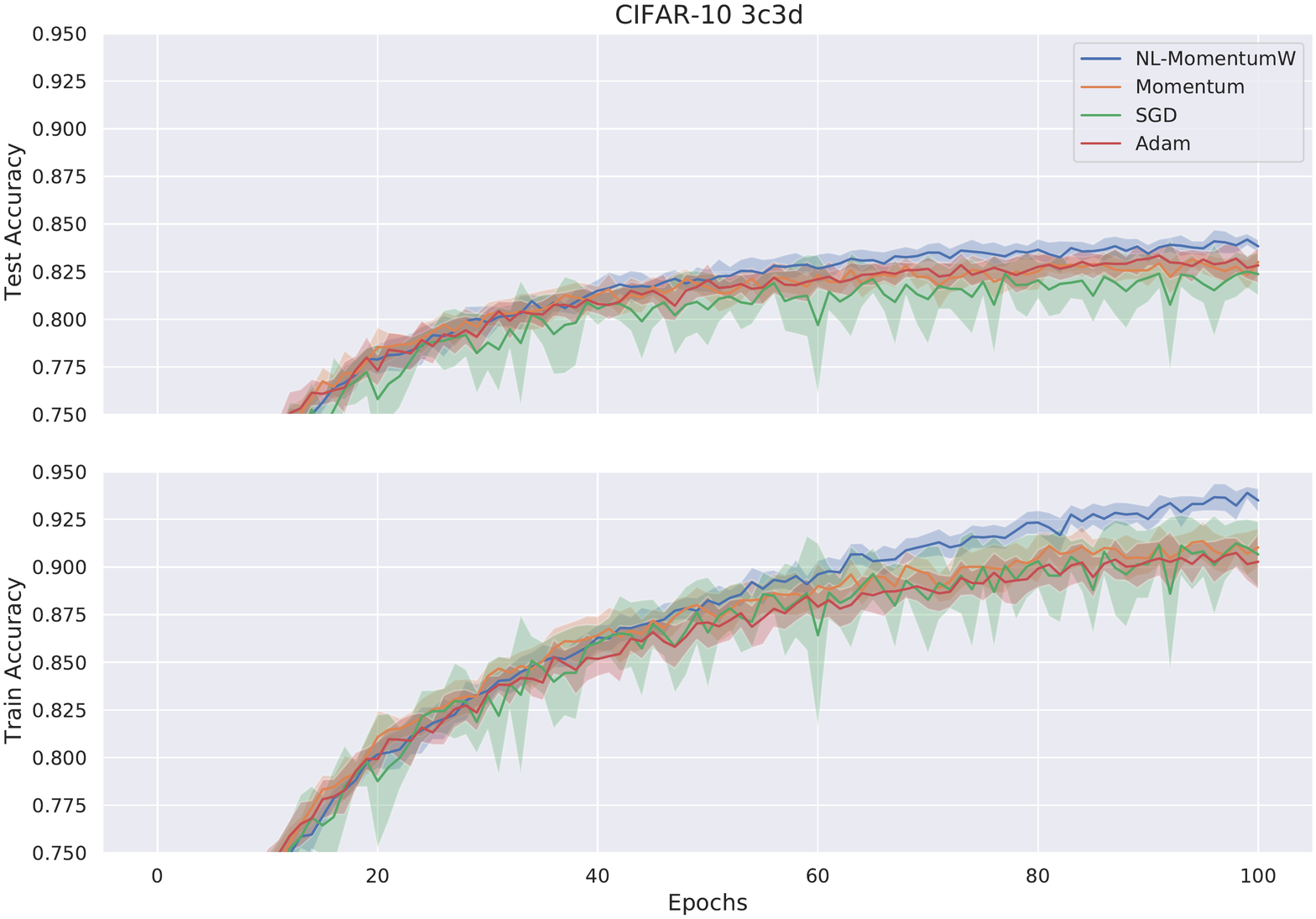}
  \caption{NL-Momentum, CIFAR-10 3C3D (accuracy)}
\end{figure*}

\begin{figure*}[!ht]
\centering
  \includegraphics[width=.7\textwidth]{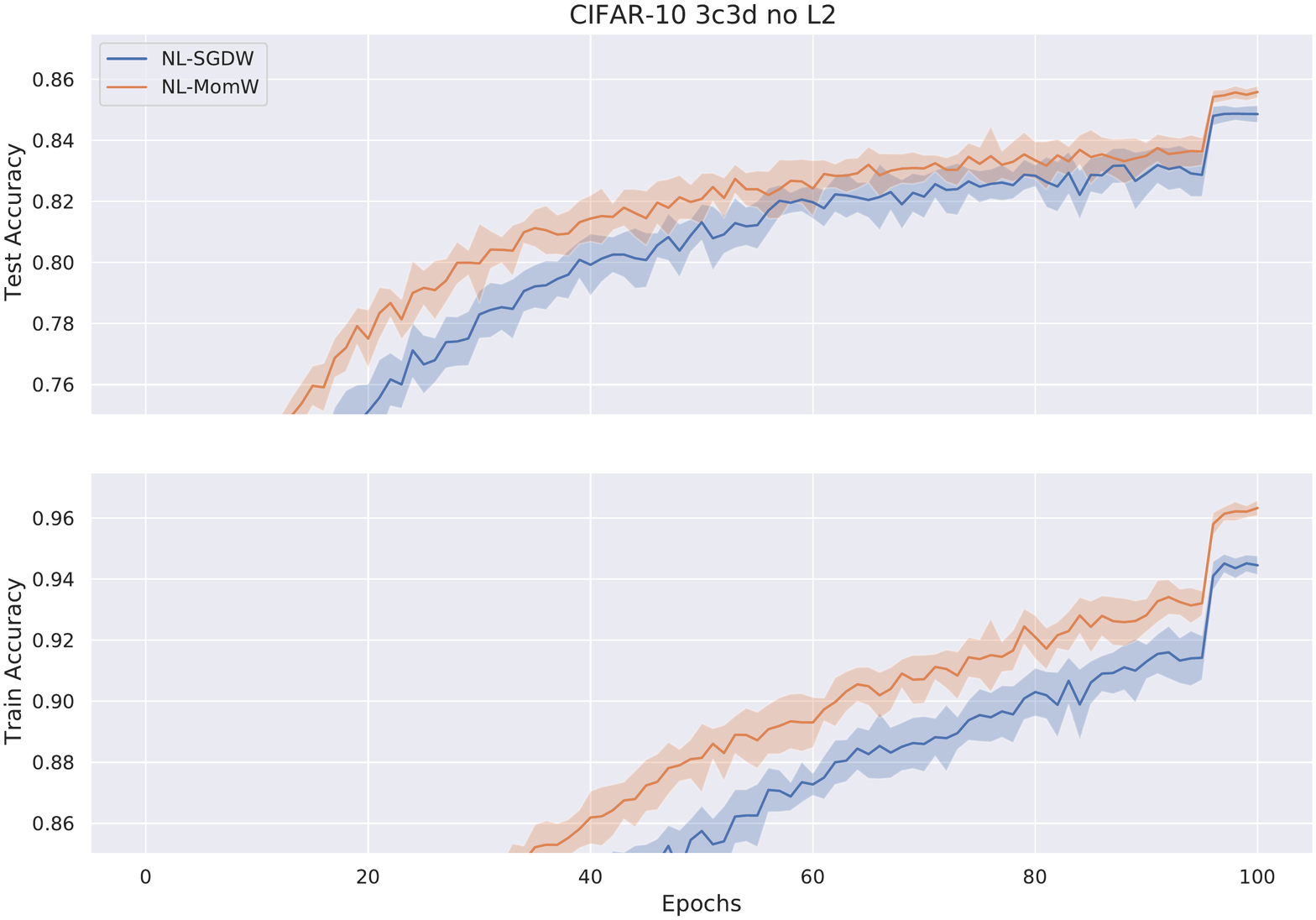}
  \caption{NL-SGD and NL-Momentum, CIFAR-10 3C3D (accuracy), effect of `annihilation' phase}
\end{figure*}

\pagebreak
\subsection{CIFAR-100 All CNNC}

\begin{figure*}[!ht]
\centering
  \includegraphics[width=.7\textwidth]{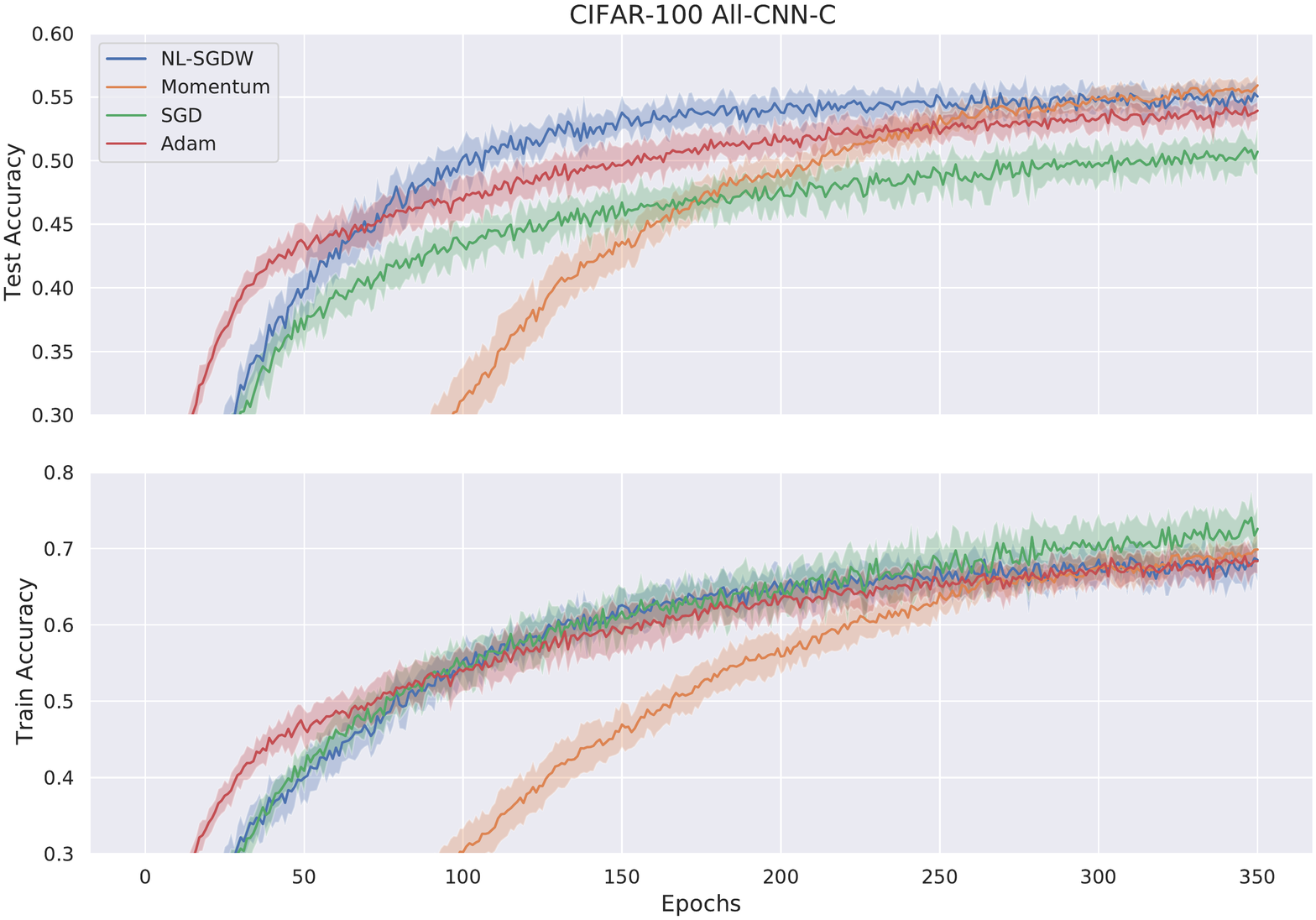}
  \caption{NL-SGD, CIFAR-100 All CNNC (accuracy)}
\end{figure*}

\begin{figure*}[!ht]
\centering
  \includegraphics[width=.7\textwidth]{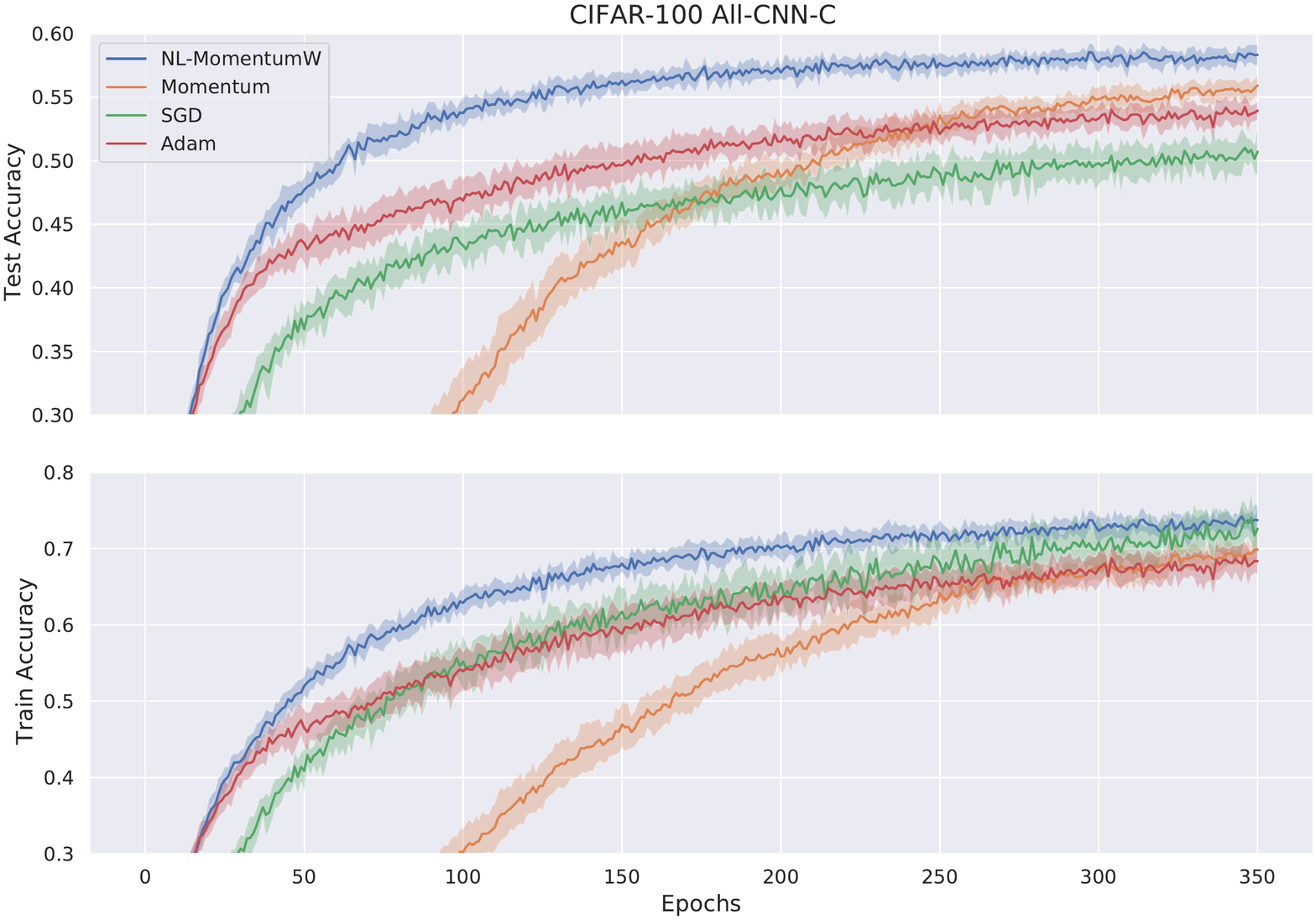}
  \caption{NL-Momentum, CIFAR-100 All CNNC (accuracy)}
\end{figure*}

\pagebreak
\subsection{SVHN WRN164}

The NL optimzers show the most interesting behaviour when applied to the SVHN WRN164 case, with accuracy improvement continuing (or even accelerating) after all the optimizers appear to have converged. If we consider the loss, we see the usual overfitting indication, but that this then levels off or, in the \nl{} case, returns to decreasing.\par

\begin{figure*}[!ht]
\centering
  \includegraphics[width=.7\textwidth]{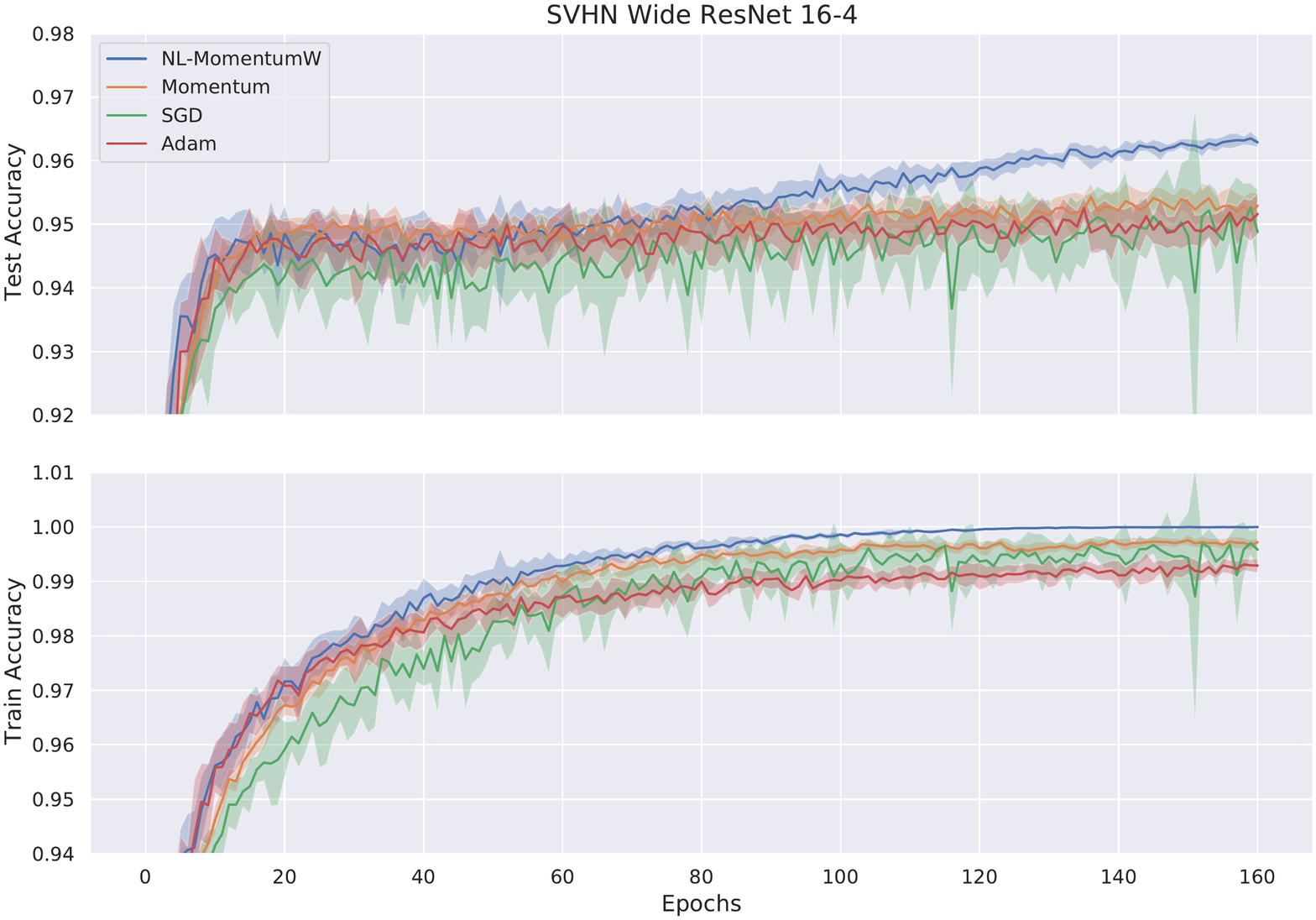}
  \caption{NL-SGD, SVHN WRN164 (accuracy)}
\end{figure*}

\begin{figure*}[!ht]
\centering
  \includegraphics[width=.7\textwidth]{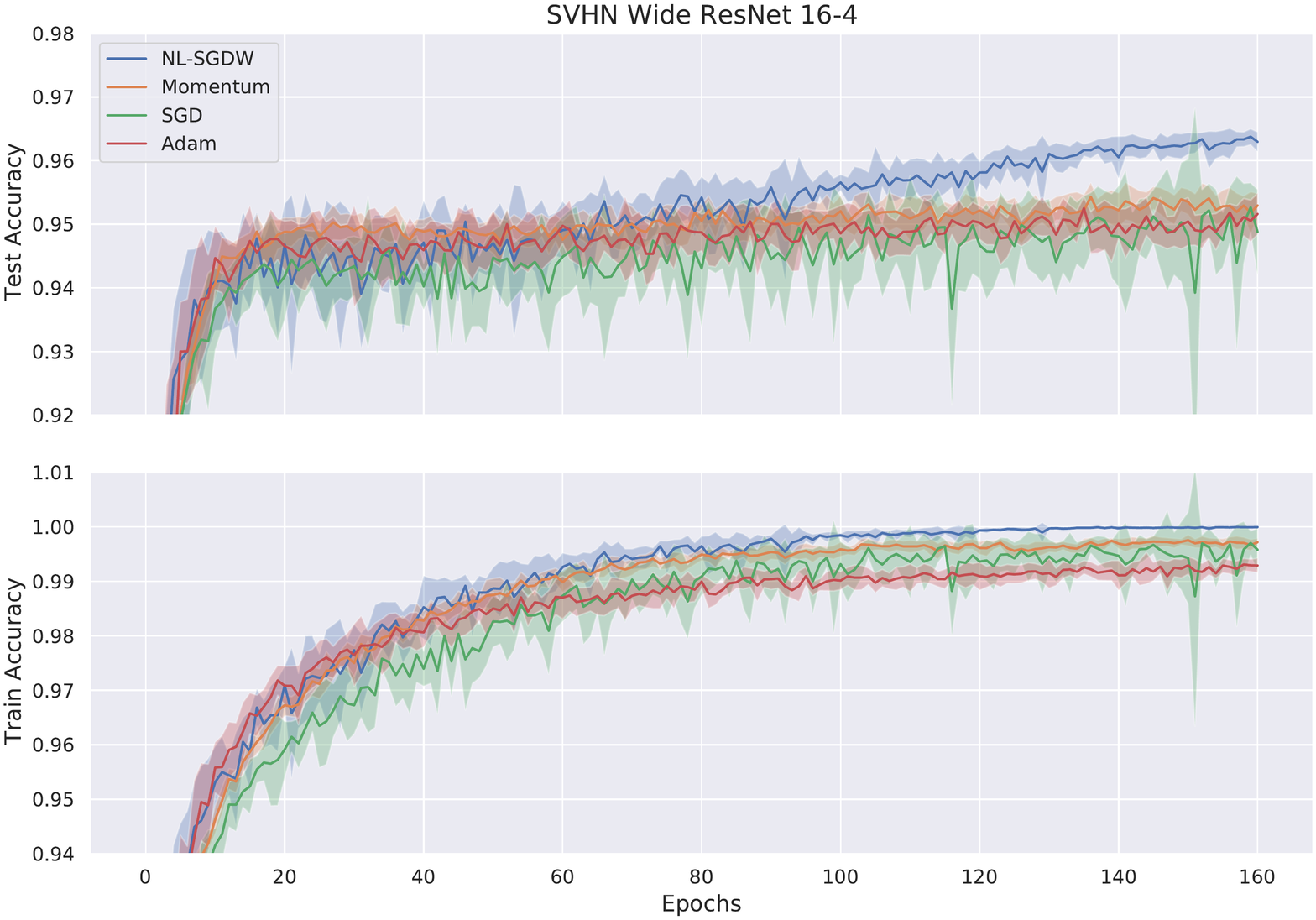}
  \caption{NL-Momentum, SVHN WRN164 (accuracy)}
\end{figure*}

\begin{figure*}[!ht]
\centering
\includegraphics[width=.7\textwidth]{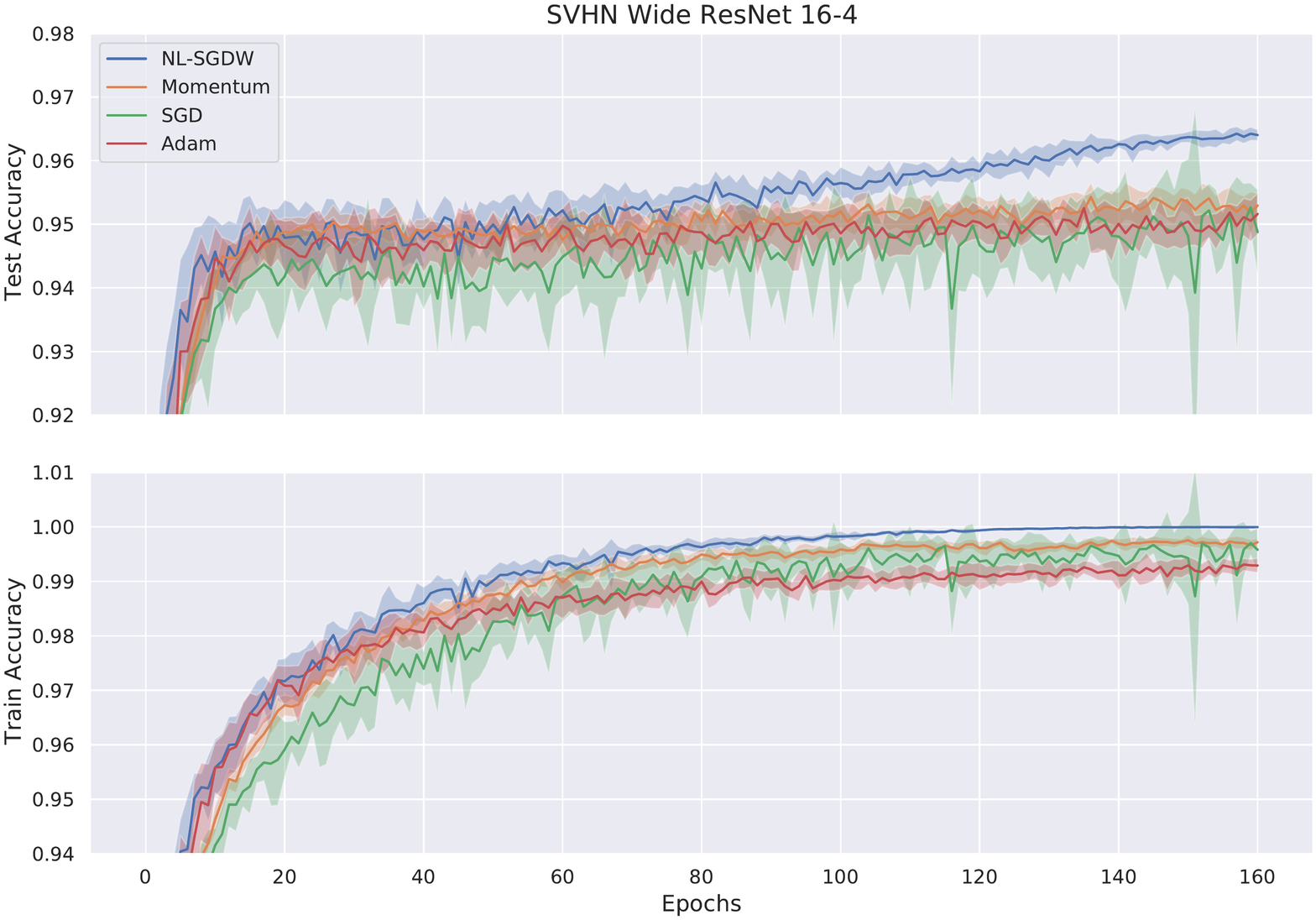}
  \caption{NL-NAG, SVHN WRN164 (accuracy)}
\end{figure*}
\begin{figure*}[!ht]
\centering
\includegraphics[width=.7\textwidth]{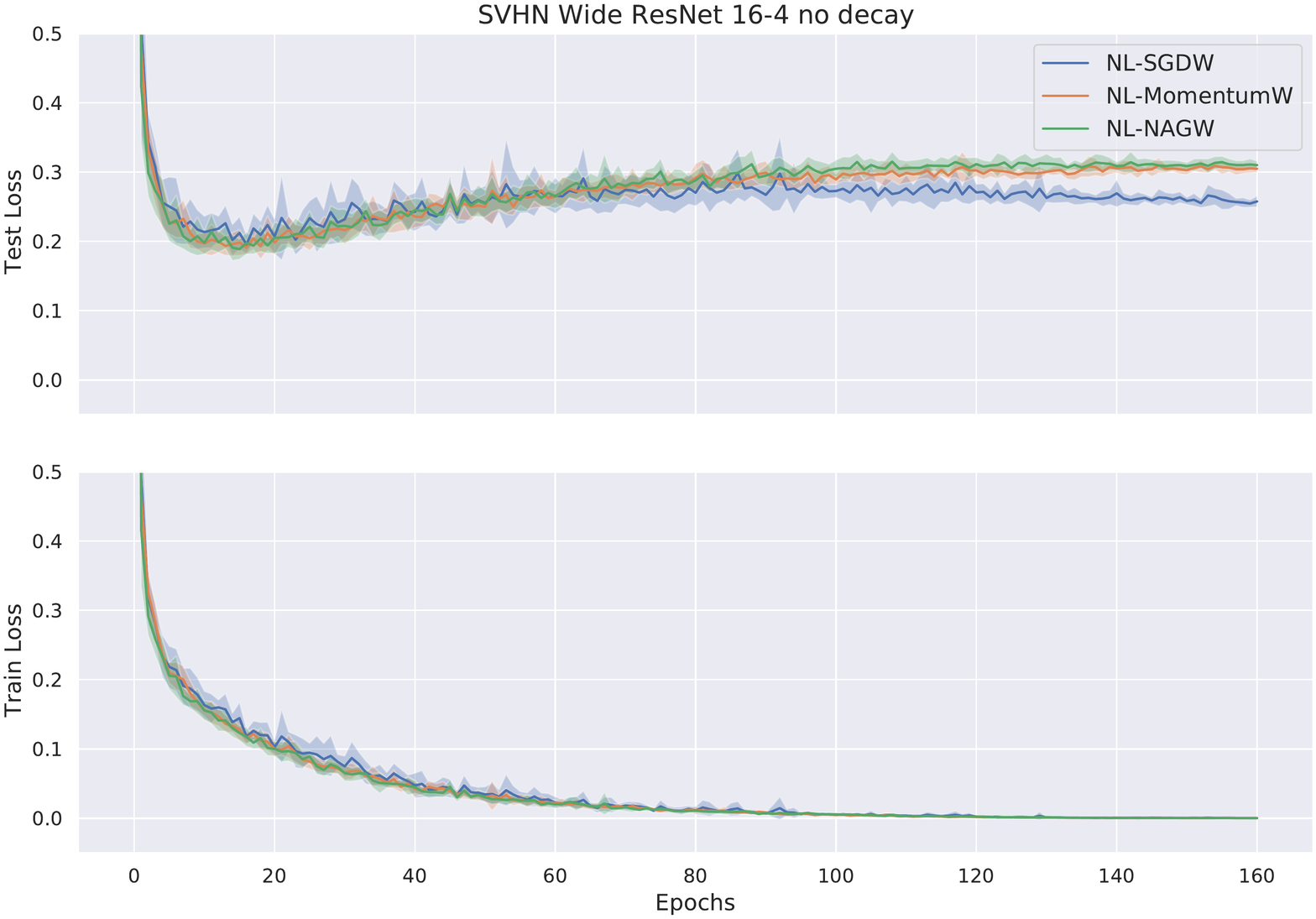}
  \caption{NL-SGD, NL-Momentum, NL-NAG (losses)}
\end{figure*}

\end{document}